\documentclass[conference]{IEEEtran}
\IEEEoverridecommandlockouts
\usepackage{amsmath,amssymb,amsfonts}
\usepackage{algorithmic}
\usepackage{graphicx}
\usepackage{textcomp}
\usepackage{hyperref}
\usepackage{natbib}
\usepackage{xcolor}
\usepackage{multirow}
\usepackage{booktabs}
\usepackage{seqsplit}

\def\BibTeX{{\rm B\kern-.05em{\sc i\kern-.025em b}\kern-.08em
    T\kern-.1667em\lower.7ex\hbox{E}\kern-.125emX}}

\begin{document}
\title{The ARIEL-CMU Systems for LoReHLT18}

\author{
\IEEEauthorblockN{Aditi Chaudhary, Siddharth Dalmia, Junjie Hu, Xinjian Li, Austin Matthews, Aldrian Obaja Muis, Naoki Otani,\\Shruti Rijhwani, Zaid Sheikh, Nidhi Vyas, Xinyi Wang, Jiateng Xie, Ruochen Xu, Chunting Zhou, Peter J. Jansen,\\Yiming Yang, Lori Levin, Florian Metze, Teruko Mitamura, David R. Mortensen, Graham Neubig, Eduard Hovy,\\Alan W Black, Jaime Carbonell}
\IEEEauthorblockA{Carnegie Mellon University\\\textit{Pittsburgh, PA 15213} }
\and
\IEEEauthorblockN{Graham V. Horwood}
\IEEEauthorblockA{Leidos, Inc.\\\textit{Reston, VA 20190}}
\and
\IEEEauthorblockN{Shabnam Tafreshi, Mona Diab}
\IEEEauthorblockA{George Washington University\\\textit{Washington, DC 20052}}
\and
\IEEEauthorblockN{Efsun S. Kayi, Noura Farra, Kathleen McKeown}
\IEEEauthorblockA{Columbia University\\\textit{New York, NY 10027} }
}

\maketitle

\begin{abstract}
This paper describes the ARIEL-CMU submissions to the Low Resource Human Language Technologies (LoReHLT) 2018 evaluations for the tasks Machine Translation (MT), Entity Discovery and Linking (EDL), and detection of Situation Frames in Text and Speech (SF Text and Speech).
\end{abstract}

\begin{IEEEkeywords}
lorelei, lorehlt, situation frames, ner, edl, mt, speech
\end{IEEEkeywords}

\section{Introduction}

The Low Resource Human Language Technologies (LoReHLT) program is a DARPA-sponsored program aiming to dramatically advance the state of computational linguistics and human language technology to enable rapid, low-cost development of capabilities for low-resource languages.%
\footnote{\url{https://www.nist.gov/itl/iad/mig/lorehlt-evaluations}}
The ARIEL-CMU team participated in three tasks (Entity Discovery and Linking, Machine Translation, and Situation Frame detection for Text and Speech) and also submitted a number of contrastive systems.
We built systems for two incident languages (ILs), IL9 Kinyarwanda, and IL10 Sinhala.


\section{Submission Highlights}


\begin{itemize}

\item NER/EDL Highlights: 
\begin{itemize}
\item For both IL9 and IL10, our NER system takes training data which are acquired primarily through cross-lingual transfer from English and related languages, and annotations from native and non-native speakers. Our system benefits from more training data, pre-trained word embeddings, gazetteers for post-processing and ensembling of models.
\item Our major improvements this year came from efficient use of both native informants and non-native annotators. Different strategies leveraging outputs from Situation Frame and Machine Translation teams were used for getting both good quality and quantity of annotations. 
\end{itemize}

\item MT Highlights:
\begin{itemize}
\item Our MT systems took a two-pronged approach of using phrase-based statistical MT and neural MT.
\item Our neural MT models were trained massively multilingually before the evaluation started, then adapted to the incident languages and other related languages.
\item We performed extensive data cleaning and selection to ensure that noise in the provided training data did not adversely affect results.
\end{itemize}

\begin{figure}[t]
\centering
\includegraphics[width=0.8\columnwidth,trim=0 2.5cm 9.125cm 0,clip]{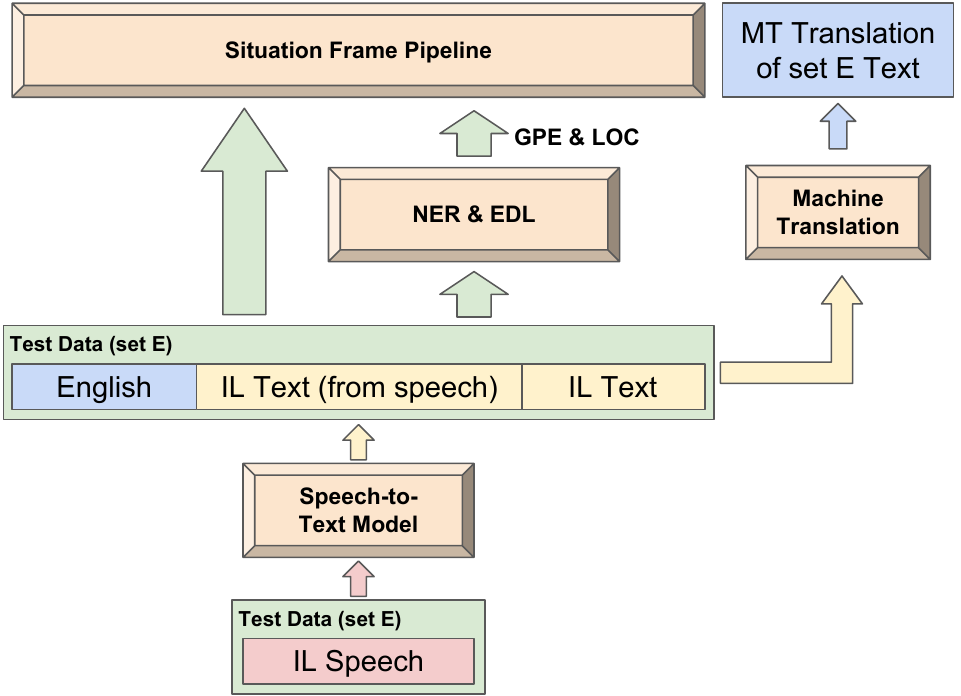}
\caption{Overall architecture of ARIEL-CMU system in LoReHLT 2018.}
\label{fig:overall-pipeline}
\end{figure}

\item SF Highlights:
\begin{itemize}
\item We processed the IL data and English data using the same models and pipelines, as our cross-lingual models can handle both IL and English texts. Speech data was first converted into text before feeding it into our SF pipeline.
\item Our major improvements this year come from training data augmentation with a bootstrapping approach and data clean up assisted with active learning.
\end{itemize}

\item SF Speech Highlights:
\begin{itemize}
\item Our Speech pipeline was designed to convert the IL speech data into text to be fed to the existing individual pipelines. This would unify the SF, NER, and EDL systems built for speech and text. 
\item Our major improvements this year came from efficient NI data collection, domain robust feature extractor, vocabulary pruning and cross-lingual transfer from Swahili for IL9.
\end{itemize}
\end{itemize}

\section{Data Resources}\label{data}


\subsection{Resources included in the IL Packs}

\subsubsection{LDC2018E55 (IL9: Kinyarwanda)}

From the LDC2018E55 pack, we made use of:

\begin{itemize}
\item the monolingual Kinyarwanda text in Set0 and Set1 (for constructing language models, word vectors, and as data for annotation and NI recording)
\item the parallel Kinyarwanda-English data in Set0 (for machine translation training data, for deriving bilingual lexicons, for training multilingual word vectors, and as an aid to non-speaker annotators)
\item the monolingual English text in SetS (for verifying our English systems and as data for annotation)
\item the included and linked Kinyarwanda-English bilingual dictionaries (for MT training data, multilingual word vectors).
\end{itemize}

\subsubsection{LDC2018E57 (IL10: Sinhala)}

From the LDC2018E55 pack, we made use of:

\begin{itemize}
\item the monolingual Sinhala text in Set0 and Set1 (for constructing language models, word vectors, and as data for annotation and NI recording)
\item the parallel Sinhala-English data in Set0 (for machine translation training data, for deriving bilingual lexicons, for training multilingual word vectors, and as an aid to non-speaker annotators)
\item the monolingual English text in SetS (for verifying our English systems and as data for annotation)
\item the included and linked Sinhala-English bilingual dictionaries (for MT training data, multilingual word vectors).
\end{itemize}

As was the case last year, the parallel text data (for both languages) had significant deficiencies. The alignment and pre-processing were both of such a poor quality that they had to be done over in order to make the data usable as training data for MT or multilingual embeddings or as scaffolding for annotators.

\subsection{Other LDC resources}
\subsubsection{LDC2017S05 (Babel Swahili Language Pack)}
This pack was used as a training set for speech recognition in IL9, mentioned in~\S\ref{kin-asr}. There are around 55k utterances in this dataset. It is the high resource language closest to Kinyarwanda.

%

\subsubsection{LDC97S44 (English Broadcast News), LDC98S74 (Spanish Broadcast News), LDC98S73 (Mandarin Broadcast News), LDC2012S06 (Turkish Broadcast News), LDC2004S01 (Czech Broadcast News)}
We used these packs for the large multilingual broadcast news speech recognition model we built for faster and better adaptation to the IL languages. The acoustics of this data is similar to the IL audio.
These packs contain around 300k utterances in total (125k from English, 100k from Turkish, 30k each from Spanish and Mandarin and around 15k from Czech).

\subsection{Additional Resources}

\subsubsection{Leidos HA/DR data}\label{sec:hadr}

As in LoReHLT17, we made use of the English in-domain text collection based on ReliefWeb, collected and annotated by our Leidos sub-team \citep{Horwood2016}. The text was used in training in-domain English language models for data selection and in creating back-translations for machine translation, and the text and annotations were used to generate keywords for our SF Keyword Model (\S\ref{model-i}) and as training data for our SF Neural Model (\S\ref{model-ii}).

\subsubsection{Leidos LRLP data}\label{sec:lrlp}

Our Leidos sub-team sampled $\sim$10K English text snippets from the LDC's LORELEI Representative Language Packs (LRLPs) and annotated them with Situation Frames, named entities, and relations among them. Snippets consist of 1-3 segments of text selected automatically according to density of terms found in the LORELEI HA/DR Lexicon. We used the English portions of parallel corpora in the following languages: Amharic, Arabic, Bengali, Chinese, Farsi, Hindi, Hungarian, Indonesian, Russian, Somali, Swahili, Tamil, Tagalog, and Yoruba. This data was used as training data for our SF Neural Model (\S\ref{model-ii}).

\subsubsection{Additional Sinhala speech data}
Read Sinhala speech data, as part SLR52~\footnote{http://www.openslr.org/52/}, SLR30~\footnote{http://www.openslr.org/30/} was used as a training set for speech recognition in IL10, mentioned in~\ref{sin-asr}. There are around 200k utterances in total combining both the datasets.

\subsubsection{Additional Swahili speech data}
Broadcast news Swahili speech data, as part of ALFFA (African Languages in the Field: speech Fundamentals and Automation)~\footnote{http://www.openslr.org/25/} was used as a training set for speech recognition in IL9, mentioned in~\S\ref{kin-asr}. There are around 12k utterances in this dataset. It is the high resource language closest to Kinyarwanda.

\subsubsection{Multilingual Bible Corpora}
Collection of Bible Audio and text aligned at a chapter level, used to create training data for speech ASR,~\S\ref{kin-asr}. It is a collection of religious texts in around 1000 languages, pre-downloaded from \texttt{bible.is}.






\subsubsection{Additional entity gazetteers}
\label{gaz}
We created gazetteers from Wikipedia resources -- Wikipedia inter-language links, and inline translations of entities in English Wikipedia articles. We also collected the high frequency n-grams from the Set1 data and translated the named entities to English, assisted by the native informants. We then manually annotated these with their respective EDL knowledge base IDs. Additionally, we compiled a gazetteer using annotations acquired from native and non-native informants.

\section{Orthography, Phonology, and Morphology}

\subsection{IPA conversion}\label{g2p}\label{ipa}

In some components, it is desirable to obtain a phonetic/phonological representation for the incident language data. This can help make the text more accessible to annotators and linguists and can reveal relationships between languages that are obscured by orthography. Data was converted from orthographic representation to the International Phonetic Alphabet (IPA) in both IL9 and IL10 using our open source G2P library, Epitran \citep{Mortensen-et-al:2018}. Epitran consists of a set of mappings between orthography and phonological representations and well as a collection of pre- and post-processors for languages where there is not a straightforward, many-to-one mapping between orthographic units and phonological units. At the beginning of the evaluation, IL9 was already supported by Epitran, but three person-hours of the first day were spent adding IL10 support. One additional hour was spent improving IL9 support.

\subsubsection{Re-romanization}

New this year was a ``re-romanizer,'' a generalized mechanism that converted IPA representations into a more familiar romanized form (similar to what is used in the romanization of foreign names in English). This meant that, while accurate IPA transcriptions were still available, annotators who were not trained in the IPA had access to a familiar representation of names in the incident languages. This was primarily of interest for IL10.

\subsubsection{Epitran backoff}
\label{epitran-backoff}

A second new addition was a backoff model, useful for mixed-language data (due to code-switching and borrowing). When using this model, the programmer instantiates Epitran with a list of language-script pairs rather than a single pair. When a token is passed to the resulting object, it attempts to transliterate as much of the token using the first language, but falls back on the other languages (in succession) when this is not possible. This is especially useful for cases, as in IL10, where documents mix scripts (Sinhala in Sinhala script and English in Latin script) and it is desirable to produce a single IPA representation of the whole document.

\subsection{Morphological parsing}\label{morphology}

For morphological parsing, we relied again on hand-crafted, rule-based systems.  However, rather than using parser combinator-based analyzers as in the previous two evaluations, we wrote morphological analyzers for Foma \citep{Hulden:2009}, a reimplementation of Xerox's XFST suite of finite state tools. These were typical Xerox-style analyzers \citep{Beesley-Karttunen:2003} with a thin Python layer providing disambiguation and a convenient interface. Using Foma allowed us to leverage existing work on the morphology of IL9 as well as create analyzers with an impressive runtime (allowing for the easy re-lemmatization of the whole dataset in short order as the analyzers improved).

At a very high level, the analyzers for the two incident languages had a very similar structure: Each consisted of an FST that parsed words having attested lemmas (the lemmas were present in the lexicon) and a second FST with a ``guesser'' that would parse forms with arbitrary stems/lemmas. For every input word, the two FSTs were applied in succession. The guesser was tried if and only if the FST for attested lemmas failed to yield a parse. Since the first FST tended to generate only a limited number of parses, this strategy helped to alleviate the problem of ambiguity that plagued our morphological parses for in LoReHLT16 and LoReHLT17.

Disambiguation was still a challenge, however, and was based on a variety of criteria. Parses were assigned costs based on the phonological shape of the lemma (e.g. a single consonant incurred a high cost for both languages) and the relative frequency of morphological properties (vocative case incurred a high cost for Sinhala). In general, our design strategy allowed for lower rates of over-parsing than those in previous evaluations.

At a low level, the two morphological analyzers were very different. The IL9 analyzer, based on an analyzer originally written for a MURI project, attempted to cover the whole morphology of the language---all parts of speech, and derivation as well as inflection. To some degree, this was suboptimal for our purposes, since most of the downstream tasks required a form of lemmatization and the ``lemmas'' output by the analyzer were actually roots rather than stems or citation forms. This likely hurt precision, while potentially benefiting recall.

The morphological analyzer for IL10 was philosophically opposite that for IL9. It represented completely original work and was tailored to the needs of the downstream tasks (except for MT). It was a very conservative parser than targeted only the inflectional morphology of nouns. It yielded lemmas that were identical to the citation form of nouns and seldom yielded more that two parses per word. When it did, the best parse was usually obvious (the one that was not nominative singular).

The morphological analyzers were used primarily to lemmatize data, which was consumed by all downstream tasks.

\section{Native Informants and Linguistic Analysts}

\subsection{Native Informants}

Data and annotations of many kinds were elicited from the native informants (NIs):
\begin{itemize}
\item \textbf{Translations}
	\begin{itemize}
    \item Translations of English SF keywords
    \item Translations of named entities (eng $\rightarrow$ IL) occurring in incident description and in set1
    \item Translations of named entities (IL $\rightarrow$ eng) occurring in Set0 and Set1
    \item Translations of high-frequency phrases and sentences in set1 
    \end{itemize}
\item \textbf{Annotations}
	\begin{itemize}
    \item EDL
    	\begin{itemize}
        \item Named entity annotations (see table~\ref{tab:ne})
        \item Correction of system output in active learning paradigm
        \end{itemize}
    \item Situation frame annotations of set0, set1, and transcribed speech
    \item Audio transcription from set0 and set1
    \item Speech recording of sentences selected from set0 and set1 by our SF system
    \end{itemize}
\end{itemize}


\subsection{Non-Native Annotators}\label{interface}

In addition to employing NIs to do translation, annotation, classification, and error correction, we made expanded use of non-speaker annotators who had varying levels of linguistic training but no direct knowledge of the incident languages. This builds upon our earlier work, in LoReHLT16 and LoReHLT17, with linguistic annotators.

We addressed the integration of annotators into our data pipelines with an improved annotation interface (based on the one introduced in LoReHLT17). This allows annotators to view multiple levels of linguistic representation, including the original text, glosses from lexical resources, IPA transcriptions, and conventional romanizations, all in one integrated and efficient interface. The same interface was used by the NIs and the annotators, but with different levels of representation visible. Using this tool, the annotators were able to produce a large number of annotations, especially named entity annotations (see table~\ref{tab:ne}).

\begin{table}
\centering

\caption{Named Entity annotations from NI and annotators by CP2}
\label{tab:ne}
\begin{tabular}{llll}
\toprule
& IL9 & IL10  & Total \\
\midrule
Tokens & 11096 & 5253 & 16349 \\
- NI & 3500 & 2569 & 6069\\
- non-NI & 7596 & 2684 & 10280\\\midrule
Total NEs (by CP1) & 891 & 567 & 1458 \\
Total NEs (by CP2) & 7570 & 2958 & 10528 \\
Unique NEs & 4231 & 2084 & 6311\\
\bottomrule
\end{tabular}
\end{table}

\subsection{Active Learning}\label{active-learning}
In order to use informants efficiently, we developed an active learning system for Named Entity Recognition (NER). This system selects sub-spans from sentences for annotation based on uncertainty under a trained model, in addition to the frequency of the tokens in the sentence. 

To compute the uncertainty, we measure the sub-span level entropy under an NER model, which is trained on the target low-resource language, by using data transferred from English and related language based on bilingual lexicons. The transfer is through word-to-word translation of the source language data. First, we cast two sets of individually trained word embeddings into a common space using a bilingual lexicon, and then retrieve the nearest neighbor target language word as the translation of the source language word. Subsequently, we train an NER model on the transferred data.

The active learning model selects the sub-spans from different sentences which have the highest entropy amongst all spans while tagging the spans with labels using the Viterbi algorithm for sequence CRF \citep{ma2016end}. The informants performed two tasks: a) annotating the the uncertain sub-spans, b) correcting the labels and the span width for the rest of the spans which were tagged using the NER model.
%
%

When the active learning could not be set up before the NI session, we select unlabeled sentences based on a heuristic method: we rank sentences by the sum of top-5 TF-IDF scores of words in it which is used to measure the importance of a sentence and meanwhile maintain the same ratio of sentence sources (WL/NW/SN) as the sentence source ratio in setE.

Additionally, we used non-native annotators to help with NER annotations by leveraging multiple linguistic resources. 
\begin{itemize}
\item Represented incident language in the IPA space.
\item Added Indicator features like honorifics (Mister, Miss, Dr, etc), location indicators (river, Mount., Mt., etc), organization indicators (Association, Ministry, etc), geo-political indicators (Republic of, etc).
\item Augmented the interface with word-by-word translations acquired from Situation Frame  and Machine Translation teams. The translations provided by Machine Translation teams were also added subsequently. 
\end{itemize}

\section{Entity Discovery and Linking}\label{edl}

\subsection{Named Entity Recognition}\label{ner}

\begin{figure}[t]
\centering
\includegraphics[width=0.95\columnwidth]{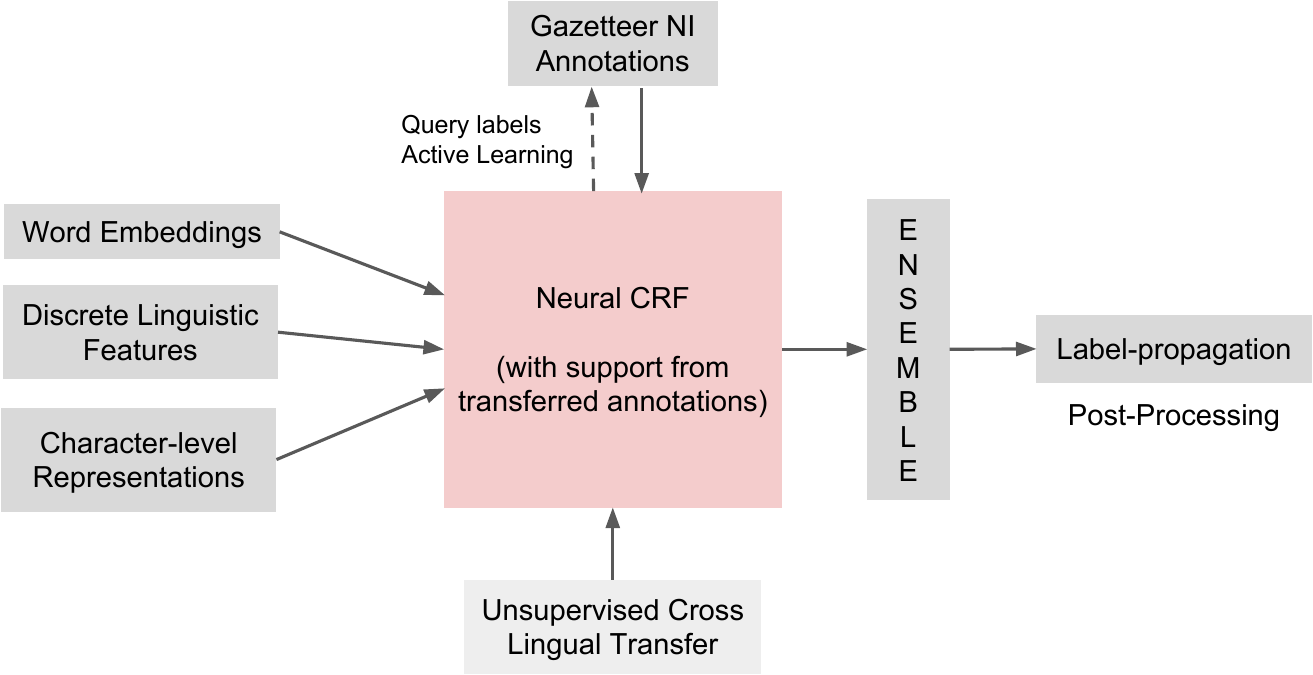}
\caption{Overall architecture of NER system LoReHLT 2018.}
\label{fig:ner-pipeline}
\end{figure}

Our NER submissions are primarily based on the neural CRF model proposed in \cite{ma2016end}, while also utilizing certain additional resources such as the compiled IL9 and IL10 gazetteer. Figure \ref{fig:ner-pipeline} provides a high level overview of our NER system.

\subsubsection{Model - Neural CRF}\label{arch}

\begin{figure}[t]
\centering
\includegraphics[width=0.9\columnwidth]{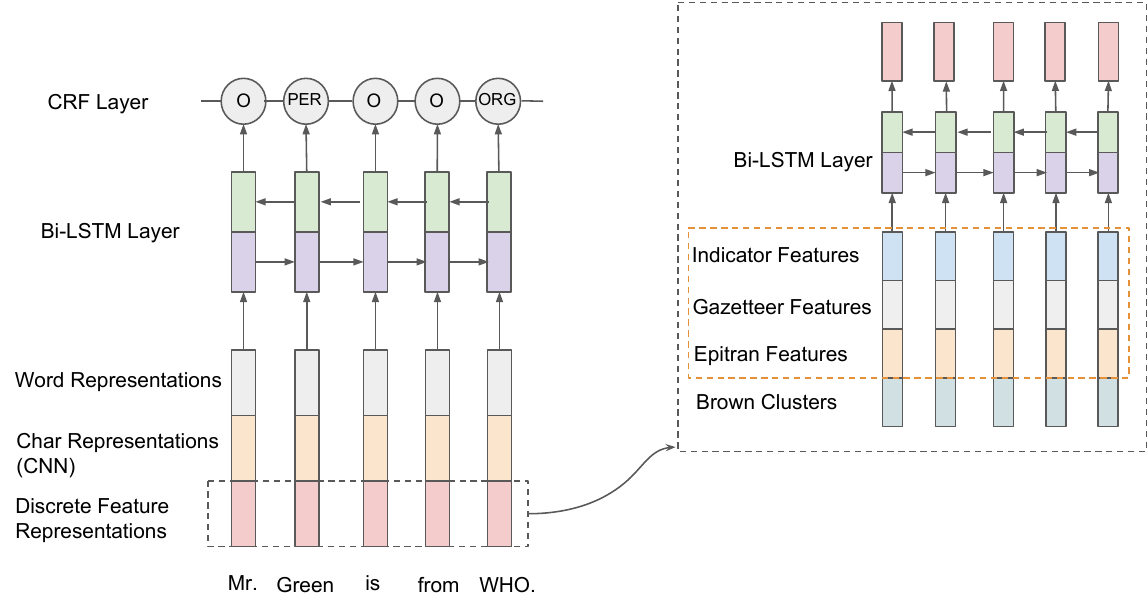}
\caption{Neural CRF architecture of NER system LoReHLT 2018.}
\label{fig:neural-crf}
\end{figure}

The neural CRF model leverages the strength of a strong neural representation learner --- words in the sequence are modeled at both type and token level. A character level convolution layer is used for modeling the token level information and is concatenated with pre-trained word embeddings which capture the type level information. FastText~\citep{bojanowski2016enriching} was used to train word embeddings for both ILs. For Checkpoint 1, monolingual data extracted from set0 and setE was combined for training; for Checkpoint 2, monolingual data extracted from set0, set1 and setE was used for training. Optionally, we also provide the provision to incorporate discrete linguistic features like indicator features, brown clusters, etc as shown in Figure \ref{fig:neural-crf}. For IL9, we notice that many entity words are capitalized words. We design the capitalization ratio feature for IL9, which is the ratio of words with capital letters (number of times word is capitalized vs total number of times appearing) over the whole monolingual corpus and we bucket this ratio to use it as one discrete feature when training models for IL9. Specifically, we calculate this ratio with the following heuristic: $( \#(word capitalized)+0.5 ) / ( \#(word) + 1.0 )$. Together, these token level representations are modeled with a bi-directional LSTM \citep{lstm}, which is known to help in tagging tasks by capturing the left and right context in a sequence.
Finally, for sequence labeling a CRF layer is used. CRF's are undirected graphical models used for calculating the conditional probability of a sequence given the observations. The use of Viterbi algorithm allows the model to perform efficient inference over the space of entire output sequences (i.e. global/sequence-level normalization as opposed to local/word level ones). 
We experimented with two strategies for combining the discrete features in the above described Neural CRF model, which we describe below. 
\begin{itemize}
\item \textbf{Model: Sep-Neural CRF} : Separate bidirectional LSTMs encoders were used to encode the embedding features (word level and character level) and the linguistic features (indicator features, brown clusters) respectively. The outputs of these two encoders are concatenated before the discriminative CRF layer.
\item \textbf{Model: Cat-Neural CRF}:
Instead of having separate encoders for different types of features, all the features are concatenated into a single continuous representation and is encoded using a single bidirectional LSTM before the discriminative CRF layer.
\end{itemize}

Submissions for both IL9 and IL10 were made by varying the different combination of features we used, for instance:
\begin{itemize}
\item Sep-Neural CRF + indicator features + brown clusters
\item Sep-Neural CRF + brown clusters
\item Cat-Neural CRF + indicator features + brown clusters
\item Cat-Neural CRF + brown clusters
\end{itemize}

\subsubsection{Noisy Training Data Acquisition}
For both IL9 and IL10, no labeled data is provided in the LoReHLT18 setting. We developed the following approaches to acquire noisy training data.
\begin{itemize}
\item \textbf{Collection of Gazetteers: }We collect named entities and their entity types from several different sources: (a) Named entities extracted from the native informants' annotation sessions; (b) Named entities extracted from non native informants' annotation sessions; (c) Name entities extracted from titles of incident language Wikipedia pages; (d) Named entities extracted from the knowledge base provided in the LDC language packages; (e) Non native informants annotated part of entities in IL-English bilingual dictionaries in the LDC IL language pack.
\item \textbf{Normalization of Gazetteer and Creation of Negative Gazetteer}: To make the collected Gazetteer above generalize to different situations, we expand it by removing special characters such as \#, @ and punctuation marks from each entity word and lower case all entities in Roman scripts. We use this normalized Gazetteer together with the original Gazetteer. For IL9, we select the top 1500 words based on the capitalization ratios described in \S\ref{arch}. We ask non native informants to pick out words that are not entities and make a negative entity set with them. For IL10, we manually make a negative entity set containing several words picked by the non native informants when they correct the Gazetteer.
\item \textbf{In-domain Data Selection for Training:} To select in-domain training data, we score each sentence in the monolingual data with the TF-IDF scores of n-grams in setE, number of key words  provided by the SF team, length of n-grams appeared in setE. For IL9, we also consider the number of words that are capitalized. We scale these scores differently to make them comparable. We rank sentences by their scores and maintain the sentence type (WL/SN/NW) ratio as in the setE to select training data.
\item \textbf{Label Propagation:} Given a Gazetteer, we use it to annotate the selected training data with label propagation as follows: we iterate over each word in a sentence and look ahead an n-gram window (from five to zero in our experiments). Once the span is found in the Gazetteer, we label the span with entity tags and skip to the next unread word. If no span is found in the Gazetteer, we label the word as a non-entity.
\item \textbf{Cross-lingual Transfer} We transfer training data from English and related languages to the target languages using bilingual lexicons. For English, we use the CoNLL 2003 training data~\citep{tjong2003introduction} and we perform transfer through word-to-word translation. First, we cast two sets of individually trained word embeddings into a common space using a bilingual lexicon, and then retrieve the nearest neighbor target language word as the translation of each source English word. For related language, we use Swahili for IL9. We use English as a pivot language to form a source-to-target lexicon as we are provided English lexicons for both languages. Different from transferring from English, we first translate source words using the resulting lexicon, then using target words with edit distance less than 1, and lastly target nearest neighbor words in the shared embedding space. The resulting training data from English and related language can then be used for IL9 and IL10.
\end{itemize}
 
\subsubsection{Target Language Specificities}
Some submissions were made specific to a particular IL which we describe in the below section.
\begin{itemize}
\item \textbf{IL9:} (a) Since IL9 has Roman script, an additional capitalization ratio feature was added as part of the discrete features. (b) For all capitalized words that are unlabeled and not in the negative Gazetteer, we mark them with UNK labels, and during training marginalize over all labels at UNK words to calculate the score of a sentence. We denote this model output ``partial-CRF''.
\item \textbf{IL10:} (a) Joint-training with Hindi was used for IL10 due to similarities in pronunciations. Word embeddings were trained by converting both IL10 and Hindi in the common IPA space. Hindi NER annotations, extracted from existing language pack \footnote{LDC2017E62} were added to the IL10 training data. (b) Edit-distance based label propagation: we didn't collect sufficient number of gazetteer items due to the fact that it is more difficult for non-native annotators to annotate non-Roman scripts. As a post-processing step, we first perform label-propagation on IL10 setE: for each word in setE if it does not exist in the Gazetteer we compare it with each word in the current Gazetteer, if they have an edit distance less than $min\_edit\_dist$, we store the entity label of the Gazetteer word. Then we assign the majority label to the unlabeled word. Empirically, we set the $min\_edit\_dist$ to be 2.
\end{itemize}

\subsubsection{Post-processing}
We first perform label propagation with Gazetteer and extract all predicted entities. Then we perform within and across document label propagation over the whole setE.

\subsubsection{English Data}
For English, we used Stanford CoreNLP~\citep{manning-EtAl:2014:P14-5} for CP1 and a vanilla neural CRF model (without additional features) for CP2. For the neural CRF model, we used data from the CoNLL 2003 dataset~\citep{tjong2003introduction} and the OntoNotes 5.0 dataset~\footnote{\url{https://catalog.ldc.upenn.edu/ldc2013t19}} for training and tuning the model. Since the data from CoNLL 2003 used different entity types, we converted them following the procedure based on Freebase type as described by \cite{TsaiMaRo16}. We used publicly available pre-trained GloVe~\citep{pennington2014glove}~\footnote{\url{https://nlp.stanford.edu/projects/glove/}} word embeddings as the word embedding inputs.

As a domain-specific preprocessing step, we performed lower-case exact string matching for all word ngrams (up to 4) of the text data against the KB (pruned as described in the next section), and tagged those found in the KB with its corresponding KB entity type. We perform matching starting from the longer ngram, and simply skip those that share overlapping spans with other named entities. We also skipped those ngrams that contained very common words, which we use the top 5,000 most frequent words in GloVe. Lastly, we tagged all hashtagged words by performing lower-case exact string matching with the hashtag and space removed and using a list of known abbreviations manually compiled from Set1 and SetS. These preprocessing steps are used to handle named entities such as rare words, lower-case words, and hashtagged words. We add more entities by running the NER system on the preprocessed texts.

Next, we detect nominal mentions using the constituency parser from Stanford CoreNLP~\citep{manning-EtAl:2014:P14-5}, which is implemented as part of our English EDL pipeline~\citep{ma-EDL:2017:TAC}. In short, we select noun phrases returned by the parsing results that do not share overlapping spans with other NEs, and perform filtering post-processing steps based on WordNet types, noun types, and etc. For more details, please refer to \cite{ma-EDL:2017:TAC}.


\subsection{Entity Linking}

\begin{figure}[t]
\centering
\includegraphics[width=0.95\columnwidth]{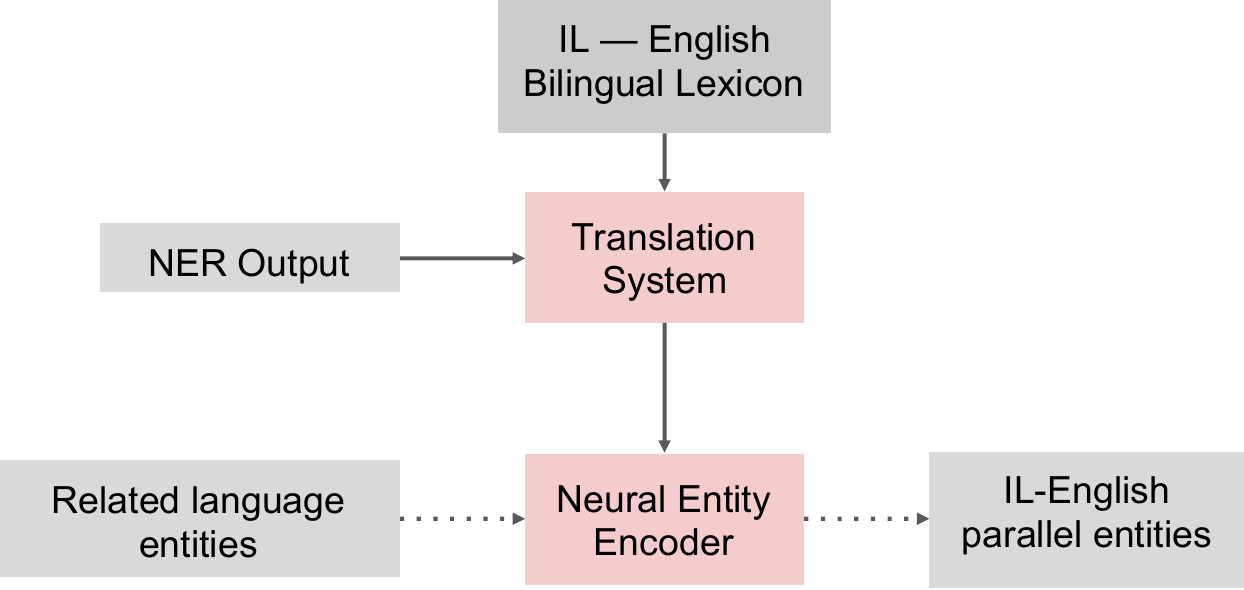}
\caption{Overall architecture of the entity linking system LoReHLT 2018.}
\label{fig:edl-pipeline}
\end{figure}

After obtaining output from the NER system, we use a entity linking methods to link the detected mentions to the knowledge base (KB). We have a high precision system that performs word-to-word translation of the mention strings and fast lookup on the KB. The mentions not linked by this system are then processed by a neural character-level encoder system. The overall architecture of the pipeline is shown in figure~\ref{fig:edl-pipeline}.

\subsubsection{Pre-Processing}
\begin{itemize}
\item \textbf{KB Pruning}: For all checkpoints, we linked mentions to a pruned version of the KB in order to reduce processing time, as well as remove entities unlikely to be related to the incident. The pruning was only for \texttt{GPE/LOC} entities, we used all the \texttt{PER} and \texttt{ORG} entities for our linking pipeline. For \texttt{GPE/LOC}, we selected all entries in the KB associated with the incident country as well as surrounding countries. We also added \texttt{GPE}s that had a population of more than 50000 according to the KB.
\end{itemize}

\subsubsection{Translation-Based Linking}
We make a first attempt at linking mention strings using a translation-based system. For each mention string, the system uses various lexical resources to generate possible English translations of the string by performing a lookup of each token in the lexicons (word-by-word translation). We then find the KB link for the mention by looking up each translation in the KB and selecting the best KB entry match according to highest Jaccard similarity on the strings, with a threshold tuned on experiments with other languages (before the evaluation).

The lexical resources we used were:
\begin{itemize}
\item Native informant translations of entities from the incident description
\item Wikipedia inter-language links (parallel article titles) between Kinyarwanda-English, Swahili-English and Sinhala-English
\item PanLex for the incident languages
\item Extracted lexicon from the parallel data in the given language packs using \texttt{fast\_align} \citep{dyer:fastalign}, pruned by the number of occurrences of the alignment.
\item Alternate names for entities in the Geonames database -- in both the incident languages as well as English
\end{itemize}

\textit{Additional entity lexicon creation}: The linguistics team and non-native annotators in the team translated over 800 high-priority entities from Set1 (300 in IL9 and 500 in IL10). Several of these were linked to the KB by team members and annotators, improving the quality of the translation-based entity linking system. A part of these annotations were used as a development set for model selection. We also asked the native informants to translate entities while annotating data for NER. A non-native annotator or EDL team member recorded these translations during the session, and mapped them back to the original IL entity as post-processing. We obtained over 300 translations for IL10 through the NI sessions. This number was less significant for IL9.

We attempted to use the morphological parsers to obtain variants of both the lexical resources as well as the mention string. However, this did not show improvement in the entity linking performance on the development set and we did not use these in our final system submission.

\subsubsection{Neural Scoring for KB Entries}

The inputs to our second linking step are entities that remain without links after running the translation-based system. This might occur because the bilingual lexicons available may not have full coverage of the input entities. We attempt to tag these entities to increase the overall recall based on character-level and phoneme-level word similarity between each of the entity and all entries in the KB. Specifically, for each entity in the input, we compute a score for each entry in the KB, and then sort the KB entries based on these scores. We set a threshold on the difference between the highest score and second highest score to determine whether the entity should be linked to the KB or remain as NIL.

For each incident language, we build a model that uses two LSTM encoders -- one that encodes strings in the incident language and the other that encodes English strings. The model is trained in order to maximize the cosine similarity between parallel entity pairs (between the IL and English). We use negative sampling with a max-margin objective during training~\citep{bordes2011learning,mikolov2013distributed}. We train models in both the orthographic (grapheme) space and the phoneme space (by converting the parallel data into IPA using Epitran \citep{Mortensen-et-al:2018}.

\textit{Joint training with related languages}: Apart from training models on the ILs themselves, we also leverage parallel data in languages closely related to the IL for training the model. Specifically, we used Swahili and Zulu for IL9, and Marathi and Hindi for IL10. For IL9, the writing systems used by the related languages are the same and we jointly trained models in both grapheme and phoneme spaces. With IL10, we used only the phoneme space for joint training. 

While testing, we use the IL encoder to encode the input mention and the English encoder to encode all the KB entries. We then compute the similarity between each KB entry encoding and the input mention encoding. The selected entity link is the top-scoring KB entry for that mention.

To determine whether an entity is NIL or linkable to the KB, we set a threshold on the cosine similarity score. This threshold is tuned on the development set created using Set1 entities. We observe that the encoder-based linking system offers diminishing utility with increasing size of the lexicon used for the translation-based entity linking system. Interestingly, with the large number of entities translated by the end of CP2, the neural encoding system offered little to no improvement in entity linking performance over the translation-based system (which offers better scores in terms of precision).

\subsubsection{English Data}

We use the EDL system of \cite{ma-EDL:2017:TAC} for the English data, which takes in English NER outputs that contain both named and nominal mentions, finds a list of candidate entities for each named entity based on string similarity, and performs document-level inference for all the named entities within the same document using graphs built from Wikipedia. The highest scoring subgraph formed by the candidate entities is selected based on a graph densification procedure from~\citep{moro2014entity}. For more details of the complete linking system, please refer to \cite{ma-EDL:2017:TAC}.

After the system outputs the entity ID for each named entity, we perform a few post-processing steps. First, as the system outputs a Wikipedia ID for each linkable named entity, we have to map the Wikipedia ID to the LORELEI KB ID. For geonames entries, we obtain the mappings for all entries that have Wikipedia links in the \textit{alternative names} table, and we perform exact string matching for all other entries to retrieve the mappings. If no mapping to LORELEI KB ID could be found for the Wikipedia ID, it would be mapped to NIL. Second, we rerun the exact string matching step of the English NER systems to handle mislinked and leftout entities, which are caused by the fact that, for example, the named entity does not exist in Wikipedia but in the LORELEI KB, or the system decides not to predict because it cannot find any reliable candidates (e.g., hashtagged words sometimes do not have high string similarity with any Wikipedia entry). Lastly, we perform NIL clustering based on the mention surface form. Note that, entities that do not share the surface form, but are linked to the same Wikipedia ID that cannot be mapped to a LORELEI KB ID, will be grouped in the same NIL cluster.

The complete list of submissions is shown in Table \ref{tab:edl_subs_il9} and Table \ref{tab:edl_subs_il10} at the end of this report.

\section{Machine Translation}

The ARIEL MT strategy was based on two pillars:
\begin{enumerate}
\item Creating a diversity of systems, as different varieties of systems work better in different situations.
\item Making best use of the resources available, including parallel IL data, lexicons, data in related languages, and monolingual data in English.
\end{enumerate}
The details of all of these methods are described below.

\subsection{System Varieties}

We created two different varieties of MT system, ones based on hierarchical phrase-based machine translation, and ones based on neural machine translation.
We also experimented with system combination to combine together multiple strong systems and achieve better results.

\subsubsection{Hierarchical Phrase-based Machine Translation Systems}\label{sec:smt}

We used hierarchical phrase-based machine translation models \citep{chiang2007hierarchical} trained using the \texttt{cdec} toolkit \citep{cdec}. We used the fast-align toolkit \citep{dyer2013simple} to do the bidirectional word alignments, and symmetrized the alignments using the \texttt{grow-diag-final-and} option. We heuristically scored parallel sentences in the set 0 data using an in-domain keyword list, and split the Set 0 data into train, development and test sets. The system parameters were tuned on the development set with MIRA for 20 iterations. Five language models were trained on Gigaword, Leidos, Set 0 English data, Set 1 English data and combination of Set 0 and Set 1 English data respectively using \texttt{KenLM}~\cite{heafield2011kenlm}. For both incident languages, we developed the morphology analyzer to extract the lemma for each source word, and trained our hierarchical phrase-based MT system on the lemmatized data.


\subsubsection{Neural Machine Translation Systems}

All neural systems were trained using the \texttt{xnmt} Toolkit \citep{neubig18xnmt} using a standard attentional encoder-decoder translation model \citep{BahdanauChoBengio2014}.
The model used one or two layers of bi-directional LSTMs \citep{lstm} for the encoder, and a single layer of LSTM for the decoder.
Training was performed with Adam \citep{kingma2014adam} with a learning rate of 0.001, batches were created based on the number of words such that the average batch size was 48 sentences, and a dropout rate of 0.3 was applied throughout the model.

The input and output were split into subword units using \texttt{sentencepiece}\footnote{\url{https://github.com/google/sentencepiece}}, using the unigram-based segmentation strategy of \citet{kudo2018subword}.
A subword vocabulary of size 8,000 was used on the target side, and 32,000 was used on the source side.

These relatively simple settings were chosen and fixed mainly because our main innovations lay in the creation and utilization of training data, rather than new neural network architectures, which we detail next.

\subsection{Data Sources and Preparation}

\subsubsection{Massively Multilingual Data Collection}

Before the evaluation, the ARIEL team gathered a large set of multilingual resources that were deemed to be potentially useful.
These resources spanned over 1,095 languages, and resulted in a total of 1.7 billion parallel sentences, spanning genres from religious texts, news, TED talks, and movie subtitles.
Many of these were gathered from the OPUS online archive \citep{tiedemann2009opus}, but also from a number of other sources.

Per the LORELEI rules, no new resources were gathered after the incident languages were announced, but the resources gathered before the eval included a small number of extra parallel resources for each incident language:
\begin{itemize}
\item \textbf{IL9:} Data from the bible (123k sentences), from the GNOME project (233k sentences), from the KDE project (39k sentences), and from the Ubuntu project (6k sentences).
\item \textbf{IL10:} Data from the GNOME project (13k sentences), from the KDE project (26k sentences), from OpenSubtitles (392k sentences), and from the Ubuntu project (6k sentences).
\end{itemize}
For each of the languages, we additionally harvested a lexicon from existing resources, as detailed in the NER section, and all native informant parallel resources were added to the training data.

\subsubsection{Data Augmentation with Entities}
One known weakness of NMT systems is that they are much worse at handling rare words, including named entities \citep{arthur2016lexicons}. This weakness is particularly problematic when we lack in-domain training data, since some name entities at test time might not occur in the training data at all. Therefore, we add synthesized training data by replacing the named entity in the training data with a random pair of named entity in the provided lexicon.  The first step for augmenting the data is to detect the  location of the named entity pairs in the parallel training data.: 1) we first run the NER tagger from NLTK\footnote{\url{https://www.nltk.org/}} on the target English side; 2) word alignments between source and target are extracted by FastAlign\footnote{\url{https://github.com/clab/fast_align}}; 3) for each named entity detected on the English side, its corresponding source side location is determined by the word alignment information. After the locations of the named entity pairs are extracted, we can easily replace each named entity pair with a randomly sampled named entity in the lexicon. 

\subsubsection{Selection of In-domain Data}
As previously stated, we collected a total of 1.7 billion sentences parallel between English and one of over a thousand other languages to be used in our polyglot neural MT system. Due to time and compute constraints we sought a smaller sub-corpus of the most relevant sentence pairs to each of the two incident reports.

To do this data selection we used a set of relevant terms extracted from setE by the NER team and compute the relevance of the English side of each sentence pair in the large corpus to these terms.\footnote{Before the LoReHLT18 evaluation period, we used terms extracted from all previous setEs, during the evaluation period we used setEs from the respective ILs.}
First, we for each English word $v$ in the vocabulary of the large corpus $\mathcal{V}$ we pre-compute the number of unique sentences in the corpus that contain $v$. We call this quantity $\text{DF}_v$ (``document frequency''), and its inverse $\frac{1}{\text{DF}_v}$ $\text{IDF}_v$ (``inverse document frequency'').

 Next, for each sentence in the large corpus we compute a (sparse) vector of length $\mathcal{V}$, where each dimension corresponds to one word $v \in \mathcal{V}$. We call the number of times the word $v$ occurs in the $i$th sentence $\text{TF}_{i,v}$ (``term frequency''). The $v$th dimension of sentence $i$'s vector is computed as $\text{TF}_{i,v} \cdot IDF_v$, the TF-IDF score \citep{salton1986introduction} of $v$ in sentence $i$. The intuition behind this approach is to have high values for words that are common in sentence $i$ (represented by a high $\text{TF}$) but are uncommon in the whole corpus (represented by a a low $\text{DF}_v$ and thus a high $\text{IDF}_v$).
 
 Given the list of relevant terms provided by the NER team, we calculate the term frequency of the word $v$ and divide again by the $\text{IDF}_v$ pre-computed on the large corpus. Finally, we rank the sentences in the large corpus by the cosine similarity between their vectors and relevant term vector, and use the highest scoring sentences (subject to some constraints discussed in \S\ref{multiling_nmt}) as our relevant sub-corpus.

\subsubsection{Data Cleaning and Filtering}

The original data provided in the language packs provided by LDC was both messy and highly misaligned.
In order to fix this problem, we performed data re-alignment, and also further filtering of sentences that did not seem to be parallel.

To do the re-alignment, 1) we concatenated all training data, and split them into different documents; 2) for each document, we split the document into small sentence segments and ran a sentence realignment algorithm using the \texttt{yasa} toolkit~\cite{yasa2013} to get the realigned data.

After re-aligning the data, we performed parallel sentence filtering based on a variant of the method described by \citet{munteanu2005improving}, as implemented in the \texttt{nafil} toolkit.\footnote{\url{https://github.com/neubig/nafil}}
This method works by training a classifier to determine whether sentences are parallel or not by taking a ``clean'' corpus where the sentences are highly parallel, and artificially introducing noise by swapping some of the neighboring sentences and labeling them as incorrectly aligned (we used a swap rate of 0.1).
This classifier is then applied to noisy data, and sentences that are labeled as incorrectly aligned are deleted from the corpus.
For IL9, we used our pre-collected version of the Bible as clean data, and for IL10 we used OpenSubtitles.
We trained a logistic regression classifier using \texttt{liblinear}\footnote{\url{https://www.csie.ntu.edu.tw/~cjlin/liblinear/}}, and removed any sentences that were deemed to be noisy with a probability of over 0.5.
As a result of filtering, the data size for IL9 reduced from 327k sentences to 296k sentences, and the data size for IL10 reduced from 434k sentences to 336k sentences.

Finally, since many of sentences in both the training data and the data we are expected to translate were extracted online from Twitter or magazines, both the source and target sentences contain a large number of identical tokens, for example, URLs, email addresses, or hash tags. We developed a tagger, called the do-not-translate (DNT) tagger, which extracts source tokens that should be simply passed through and not translated into English.
These tags are removed before translation, and restored after translation.

\subsection{Multilingual Training of NMT}\label{multiling_nmt}

To take advantage of the large-scale multi-lingual resources, we performed multi-lingual training of our neural machine translation systems.

Before the evaluation started, we trained a large system from all the 1,095 languages in our database into English. This system was trained on data chosen such that the threshold of the TF-IDF training criterion above was greater than -9, and at least 4,000 sentences were included per language. This resulted in a total of approximately 60M sentences in the training data.

Once the evaluation started, we started adapting this pre-trained system to the incident languages.
This was done by taking the pre-trained model and re-initializing only its word embeddings to reflect the new vocabulary in the source language, then continuing training of all parameters of the model.

In addition to performing this continued training on only the incident language itself, we also tested models that performed training with the source language and related languages, again using the TF-IDF based data selection to select relevant data.
Specifically,
\begin{itemize}
\item \textbf{IL9:} We were not able to find large amounts of data for any of the typologically related languages for Kinyarwanda.
However, because both Kinyarwanda and English are written in Roman script, and because many of the entity names are shared, we decided to add additional English-to-English data as a pseudo-translation task.
This data was selected so that the TF-IDF threshold was greater than -7, resulting in 317k sentences that contained keywords related to the incident.
\item \textbf{IL10:} For Sinhala, we were fortunate to have reasonably sized resources for two related languages: Hindi and Bengali.
We used all of the resources in our database for these two related languages, which resulted in a total of 4.39M training sentences.
\end{itemize}

\subsection{System Combination}

For the final submission, multiple systems were ensembled together to create the final results.

For combining NMT systems that share identical output vocabularies, it is simple to perform ensembling at hypothesis generation time, where multiple systems are run in parallel, and the average of the predicted word probabilities are used to predict the next word in the sequence \citep{sutskever2014sequence}.
We used this method to combine together multiple NMT systems.

In addition, to combine more heterogenous systems, we used the MEMT \citep{heafield2010combining} system combination method.
Since MEMT requires a large $n$-gram language model to rescore hypotheses, we built a large 4-gram Kneser-Ney \citep{kneser1995improved} language model using KenLM \citep{Heafield-estimate}.
We then combined the 1-best output of eleven (for IL9) or seven (for IL10) of our best neural MT systems. Training was performed using setF and all of the MEMT toolkit's default settings.

\subsection{Creation of Data w/ Native Informant}

The native informant sessions were used to create two varieties of data: word or phrase lexicons, and translated sentences from set1.

To select data words or phrases for the native informant to translate, we used the method of \citet{bloodgood2010bucking}, which selects words or phrases that occur in monolingual data (i.e. set1), but not in bilingual data (i.e. set0), sorted in descending order of frequency.
We additionally follow \citet{miura-EtAl:2016:N16-1} in removing shorter phrases that are completely subsumed by longer phrases.
These words or phrases were translated by having the native informant type translations in a Google Sheet.
This resulted in approximately 200 high-frequency words/phrases in each language.

We also translated sentences from each of the three genres of text: newswire, social networks, and weblogs, to be used as development data to assess the accuracy of our systems.
These sentences were translated by having the native informant speak the translations out loud, while a member of the ARIEL team typed in the English translations.
This resulted in 79 sentences in Kinyarwanda, and 233 sentences in Sinhala.

\subsection{Final Submitted Systems}

Our NMT system submissions fall into several categories: 1) NMT-adapt: we pre-train a large NMT system on the multilingual corpus collected as described previously, then fine-tune the system on the incident language; 2) NMT-mult: the NMT system is trained from scratch on a concatenation of the incident language training data, and the parallel data of two related languages in our multi-lingual corpus. For Kinyarwanda, we used Swahili and Zulu. For Sinhala, we used Marathi and Hindi; 3) NMT-plain: the NMT system is trained from scratch on the incident language training data only.

Here we provide a summary of our submissions. The complete list of submissions is shown in Table \ref{tab:mt_subs_il9} and Table \ref{tab:mt_subs_il10} at the end of this report.

\subsubsection{Checkpoint 1}
The submission statistics of different systems is summarized in Table \ref{tab:cp1_stats}.

\paragraph{SMT} We trained the hierarchical phrase-based system described in \S\ref{sec:smt} on the LDC tokenized data. We also made submissions that utilized the realigned data. 

\paragraph{NMT} For checkpoint 1, a particular challenge for utilizing multilingual corpus by NMT-mult is that training takes much longer time to converge because
of increased amount of training data. NMT-plain takes much less time to train than NMT-mult, but its performance might not be as good as models that utilize multi-lingual corpus.  On the other hand, NMT-adapt can quickly adapt to the incident language while taking advantage of the multi-lingual corpus. Because of the time constraint, we had NMT-adapt submissions for both languages and other NMT systems for only one language. 

\begin{table}
 \small
  \centering
  
  \caption{Checkpoint 1 MT submission statistics (all constrained)}
  \label{tab:cp1_stats} 
  \begin{tabular}{l|c|ccc}
  \toprule
  \multirow{2}{*}{\textbf{Model}} & \multirow{2}{*}{\textbf{SMT}} & \multicolumn{3}{c}{\textbf{NMT}} \\
   & & adapt & mult & plain \\
  \midrule
 IL9 & 4 & 2 & 1 & 3 \\
 IL10 & 1 & 2 & - & - \\
  \bottomrule
  \end{tabular}
\end{table}

\subsubsection{Checkpoint 2}
The submission statistics of different systems is summarized in Table \ref{tab:cp2_stats}. For checkpoint 2, we performed data filtering for realigned data to further remove misaligned data. We also used the small parallel data of set1 created with the help of native informant for system evaluation.  


\paragraph{SMT} A big challenge for the hierachical phrased-based MT system is that the system cannot translate source words that are morphological variants of their lemmas. Some of our attempts include: 1) we made submissions that utilizes words segmented into morphemes. 2) we tried to further split words into subword units by the \texttt{sentencepiece} toolkit.

\paragraph{NMT} We found that in general, the NMT-adapt outperformed the NMT-mult and NMT-plain by a large margin on the small set1 test set we created, so we focused on tuning NMT-adapt for checkpoint 2. Some of our attempts include: 1) we added back-translation data by directly copying monolingual English data as the source. This was especially helpful for Kinyarwanda, as it encouraged the model to pass through the English words on the source side; 2) we averaged system checkpoints for decoding; 3) we ran system combination on all the NMT outputs.

\begin{table}
 \small
  \centering
  
  \caption{Checkpoint 2 MT submission statistics}
  \label{tab:cp2_stats}
  \begin{tabular}{l|c|cc|c|cc}
  \toprule
   & \multicolumn{3}{c|}{\textbf{Constrained}} & \multicolumn{3}{c}{\textbf{Unconstrained}} \\
  \midrule
   \multirow{2}{*}{\textbf{Model}}& \multirow{2}{*}{\textbf{SMT}} & \multicolumn{2}{c|}{\textbf{NMT}} & \multirow{2}{*}{\textbf{SMT}} & \multicolumn{2}{c}{\textbf{NMT}}  \\
   & & adapt & mult & & adapt & mult\\
  \midrule
 IL9 & 4 & 6 & - & 4 & 5 & 1 \\
 IL10 & 3 & 7 & - & 2 & 7 & 1\\
  \bottomrule
  \end{tabular}
\end{table}

\section{Situation Frames}\label{sfs}
In this year's evaluation, we used the same SF type classification pipeline for both text and audio input by first converting the audio files into text.

We note that all of our systems satisfy the constrained condition of the task, and in our submissions we marked 10 of them as constrained based on what we believe to be the best systems.
The summary of our submissions with different settings are shown in Table \ref{tab:cp2_sf_stats}. KW refers to our Keyword Model, and NN refers to our Neural Model. Each submission consists of English, IL text, and IL speech output.
The complete list of submissions is shown in Table \ref{tab:sf_subs_il9} and Table \ref{tab:sf_subs_il10} at the end of this report.

  

\begin{table}
 \small
  \centering
  
  \caption{Number of SF submissions in various settings}
  \label{tab:cp2_sf_stats}
  \begin{tabular}{l|c|c|c|c}
  \toprule
   & \multicolumn{2}{c|}{\textbf{Constrained}} & \multicolumn{2}{c}{\textbf{Unconstrained}} \\
  \midrule
   {\bf Model:} & {\bf KW} & {\bf NN} & {\bf KW} & {\bf NN} \\
  \midrule
  \multicolumn{5}{c}{\sc Checkpoint 1} \\\midrule
 IL9 & 6 & - & - & -\\
 IL10 & 7 & 3 & 4 & -\\\midrule
  \multicolumn{5}{c}{\sc Checkpoint 2} \\\midrule
 IL9  & 4 & 6 & 5 & 5\\
 IL10 & 5 & 5 & 5 & 5\\
  \bottomrule
  \end{tabular}
\end{table}

\subsection{Speech to Text}\label{speech}
The main focus of our Speech SF pipeline was on making decisions on grounded words which would unify the pipelines of all the tasks. This involved building an Automatic Speech Recognition~(ASR) system for both the languages. Sinhala being a higher resourced language compared to Kinyarwanda in terms of resources that we had for the ASR, led us to apply different approaches for each of the language. The core model of our speech recognizer remained the same as last year. We used a sequence based ASR using CMU's EESEN system~\citep{miao2015eesen} trained using the Connection Temporal Classification~(CTC) loss. The target labels were generated using Epitran~\citep{Mortensen-et-al:2018} the grapheme to phoneme library discussed earlier in this report. It was used to generate lexicons for the words present in the training, development and test set. To ground the acoustic model output to words a WFST based decoding was done using a language model that was built on the monolingual text corpus~(lowest perplexity language models was chosen between 3-gram and 4-gram models). The decoding vocabulary was carefully chosen ensuring that we don't miss any possible locations and situation frame keywords. 

\begin{figure}[t]
\centering
\includegraphics[width=0.85\columnwidth,trim=0 2.5cm 14.5cm 0,clip]{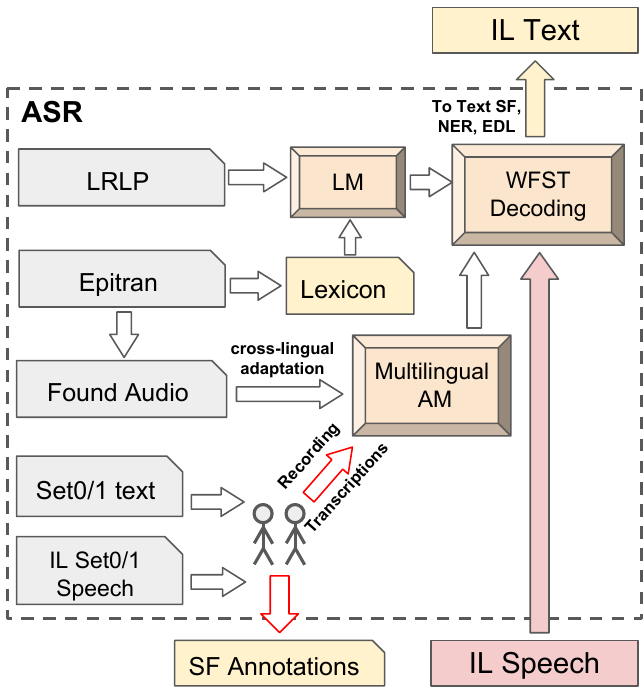}
\caption{Speech Pipeline of ARIEL-CMU system in LoReHLT 2018.}
\label{fig:speech-pipeline}
\end{figure}

To perform ASR on the IL speech data provided to us we reduce the silence and split the data into small single speaker segments using 4-class Hidden Markov Model~(HMM) segmentation followed by BIC clustering, based on the lium toolkit~\citep{meignier2010lium}. This automatically segmented incident data was converted to text using the ASR and passed on to the Text SF, NER and EDL models. An illustration of our pipeline is shown in Figure~\ref{fig:speech-pipeline}.

\subsubsection{Speech Recognition System}
\paragraph{IL9 Speech Recognition System}\label{kin-asr}

For CP1, due to the short development time, we had almost zero training data available. We tried developing a system on our Kinyarwanda and Kirundi Bible data. This was aligned automatically using a speech synthesis module, explained in \S\ref{tts}, which was later fine-tuned using downstream and upstream tasks of ASR and speech synthesis system~\citep{prahallad2007automatic}. The bible ASR was trained using the domain robust features~\citep{dalmia2018domain}. The bible ASR output was grounded to words using a beam search decoding using the ASR and phoneme RNN based Language Model. 

We found that our Kinyarwanda output for CP1 was not so good and after some more verification using the NI collected data we figured that the alignment between the text and audio of the Bible data was not reliable enough to train a system. 

For CP2, we shifted our focus to trying multilingual models and tried to transfer models from a close high resource language, discussed in ~\citep{dalmia2018sequence}. We used a pre-trained multilingual broadcast news model of English, Turkish, Spanish, Czech and Mandarin, using the resources mentioned in \S\ref{data}, which we adapted to Kinyarwanda using some phoneme mappings. Even though this gave us improvements over our CP1 model, the languages being used to transfer were very far away from the incident language. After some careful selections we found that Swahili was the closest language to transfer the ASR from. For our final model we trained a Swahili based recognizer, using the resources mentioned in \S\ref{data} and mapped its phones to Kinyarwanda. We also added around 550 utterances collected from the NI. This was crucial to improve the recognizer and fix some of the phonemes confusions that had occurred due to transfer from Swahili. To ground ASR to words we used WFST based decoding. 

We cleaned the monolingual newswire text for the language model training where we filtered out all scripts except Latin. The vocabulary of the Kinyarwanda decoder was restricted to around 100k.

\paragraph{IL10 Speech Recognition System}\label{sin-asr}
For CP1, since we had available Sinhala speech data (\S\ref{data}), we started developing an ASR directly on that. Around 180k utterances were chosen to train the model and 1.5k was used as the validation set. We built a WFST based decoding graph using a trigram language model of set0 and setE data for checkpoint one. The best decoding parameters were chosen based on the performance of the system on NI recordings.

We found that there was a clear mismatch between the training data and the IL audio, which was mostly broadcast news data. Which could potentially lower the quality of the ASR.

For CP2, we used the domain robust feature extraction technique discussed in ~\citep{dalmia2018domain}. This gave us an 10\% relative reduction in WER in the NI collected data. We also improved our language model by using more in-domain set1 data. We restricted our model to only newswire text. To clean the monolingual newswire text for the language model we filtered out all scripts except Sinhala and English. Sinhala being a non-Latin script we assumed Sinhala to contain Latin loan words and be influenced by English. The vocabulary of the Sinhala decoder was restricted to around 50k. 

\subsubsection{NI speech annotation systems}
To collect speech data from NI effectively, we developed two web applications to interact with NIs. During the NI sessions, we sent links of those applications to NIs and they followed instructions to collect speech data.
\begin{enumerate}
\item The first application is the annotation application in which we can collect transcriptions of specific audios we present to NIs. During the testing period, we found that it was much easier for NI to transcribe shorter audio clips, because longer audios usually contains a lot words that they requires them to replay the clip several times. Prior to the NI session, we automatically segmented speech audio using the technique mentioned in the previous section. All clips shorter than 2 seconds were ignored as they would usually contain background noises or music. Clips between 2 seconds and 6 seconds were played to the NI after being manually verified to see if the automatic segmentation did not miss out any clips containing music or unrelated noise. If the NI thought the text being spoken in the audio could clearly define an SF type then that was noted and passed to the text SF as development data.
\item The second application is the recording application. This application allows NI to record their voice by reading specific texts we have prepared. The recording application is more effective than the previous application in terms of collection speed as it is easier to read sentences rather than transcribe audios. It is particularly useful when NI is poor at typing or transcribing audios. However, the major drawback of this application is that speakers are confined to the few NIs, and their recording environment does not match audios in our datasets. Additionally, some background noises in NI's room also makes the recording harder. The text that the NI was asked to read was set0/set1 segments of documents that the SF keyword system thought was high confidence. This way we could even verify if the SF prediction was correct.
\end{enumerate}

During the entire sessions with NIs, we collected 680 audio/transcription pairs for IL9 and 477 audio/transcription pairs for IL10. As IL9 had very less speech training data, we used most of them as the training set and reserved 100 utterances for the validation purpose. On the other hand, IL10 has sufficient amount of training data and we used all utterances as the validation set.

\begin{figure*}[ht!]
\centering
\includegraphics[width=0.8\textwidth,trim=0 1.5cm 0 0,clip]{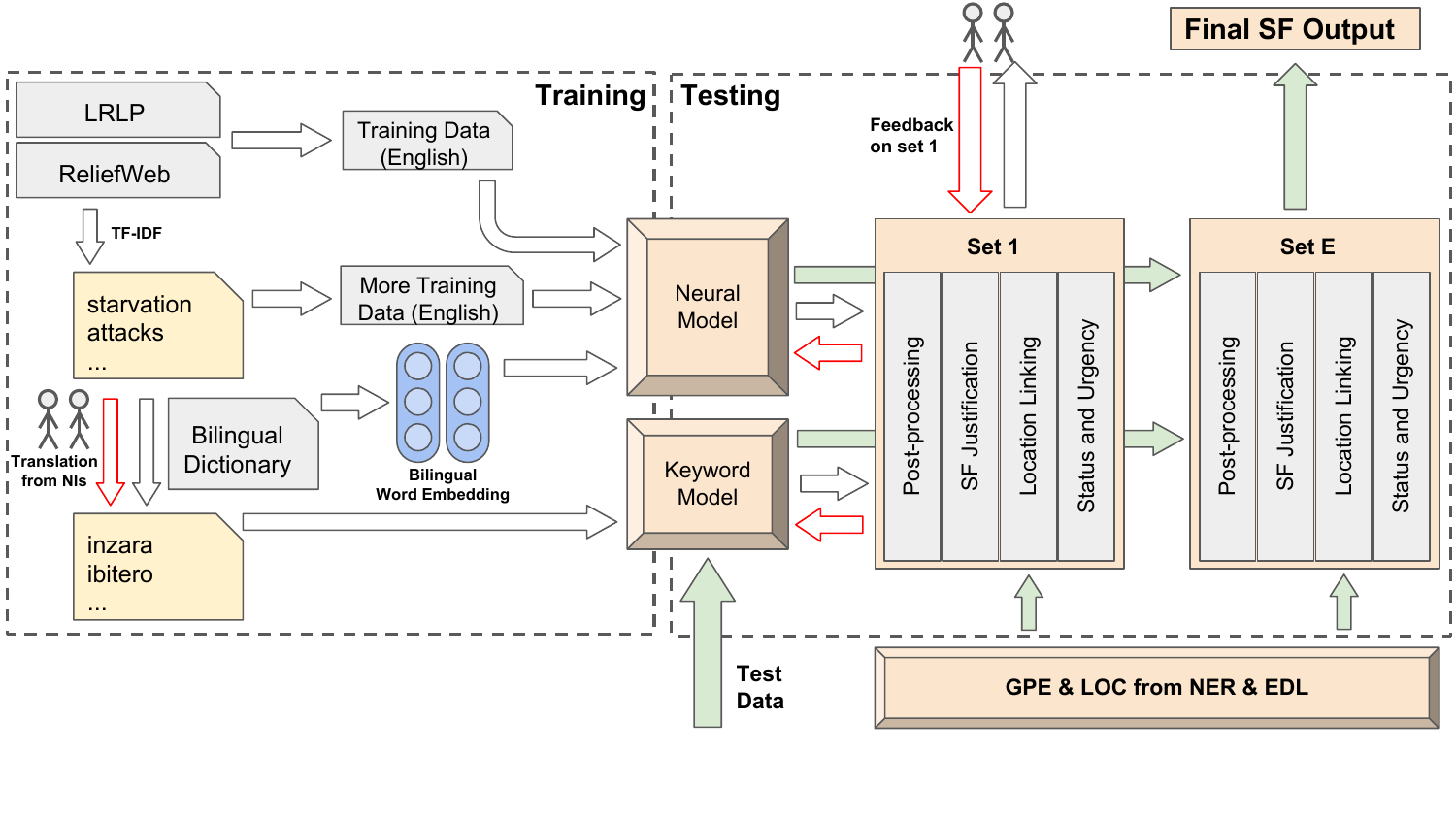}
\caption{Situation Frame identification pipeline, which includes SF Type, SF Location, SF Justification, Status, and Urgency.}
\label{fig:sf-pipeline}
\end{figure*}

\subsubsection{IPA conversion for Speech}\label{sec:ipa-speech}
We performed all of the experiments in the IPA space. This is particularly useful because it makes it easy to transfer acoustic models across languages. We found high overlap of Kinyarwanda and Swahili phonemes which we believe was crucial to make the transfer successful. Even though we grounded the words back to its orthography as part of the WFST decoder. We converted the orthography into IPA using the Backoff mode in Epitran (\S\ref{epitran-backoff}), which helped filter out noise in the IL text like urls, non IL scripts, emoticons, numbers. These usually tend to effect the decoding of the ASR. In our initial experiments we also found that apart from the filtering of text, doing SF in the IPA space could be useful if there is high irregularity in the orthography and can often help in normalizing some of the spelling errors.

\subsubsection{Speech Synthesis}\label{tts}

We built speech synthesizers for Kinyarwanda and Kirundi. We used CMU's Clustergen Parametric Speech Synthesis system \citep{black2015random} as it is robust to data noise and produces reliable synthesis even for small amounts of data, pronunciations were produced by Epitran.  We further used these synthesis models to align read Bible data in the incident language (even though that data had background music) to produce synthesizers with actual native acoustic data \citep{prahallad2007automatic}.

\subsection{Situation Frame Pipeline}
The identification of SFs and selection of SF types is performed by two primary models, each with numerous variations. The sentences (and the surrounding context sentences) justifying the models' predictions are used to further enrich the situation frames with location, status, resolution, and urgency. An illustration of our pipeline is shown in Figure \ref{fig:sf-pipeline}.

\subsubsection{SF Type Identification}

\paragraph{Keyword Model}\label{model-i}
Keyword Model is a lexicon matching model using a list of curated keywords for each SF type. We created the list of keywords in two steps: (1) build a list of keywords for each SF type in English, then (2) translate the keywords into the target language by using the provided dictionary and also by the native informants during NI sessions in the first checkpoint. During the translation process by the NIs, the NIs were shown English keywords with an example usage taken from ReliefWeb. The idea is to provide the NIs with the context of the English keywords in order to get the most relevant translation.

Building English keywords (step 1 above) is a two-step process. First, we used the ReliefWeb dataset to generate a list of 100 candidate keywords for each class by taking the top-100 words with the highest TF-IDF scores. Similar to the keyword generation method described by~\citet{Littell2017}, we manually refined the keyword list by pruning based on world knowledge. For each candidate keyword, we added 30 most similar words using the English word2vec model trained on the Google News corpus.\footnote{\url{https://code.google.com/archive/p/word2vec}} We retained only those words which have cosine-similarity scores greater than 70\%. For each candidate keyword in this extended list, we computed a label affinity score with each class label~(e.g., {\it water}, {\it evacuation}) using cosine-similarity between their word2vec embeddings. Candidate keywords with similarity above a certain threshold $th_1$ were retained and are used as keywords for the corresponding classes. We primarily used $th_1=0.8$ in our submissions, but also submitted versions with $th_1=0.9$ for comparison.

During testing, we retrieved sentences that contain any of the keywords in our list, and assigned the top-2 SF types into the sentences that contain the respective keywords, based on the sum of confidence scores of the keywords.

At checkpoint 2, we used NI sessions to verify the outputs of the Keyword Model, and used the results to prune certain keywords which are not useful for prediction, based on our interactions with the NIs. By this time, we also used a morphological analyzer from our linguistics team for each IL for matching lemmatized keywords in order to improve recall. We think this will be particularly useful for IL9, which is morphologically rich.

We also made another submission where we did the keyword matching on the IPA version of the IL texts and of our speech-to-text outputs. The keywords were first converted into IPA, and matching was done as usual.

\paragraph{Neural Model}\label{model-ii}
Our second model is a convolutional neural network (CNN) that takes sequences of word embeddings as input and classifies them into SF types.

The first step is to train a bilingual word embedding as a shared feature representation between English and ILs.
We used XlingualEmb~\citep{duong2016learning} and trained our bilingual word embedding for English and the IL. XlingualEmb is a cross-lingual extension from word2vec model \citep{mikolov2013distributed} to bilingual text using monolingual corpora and a bilingual dictionary.

Then the CNN model takes a sequence of (bilingual) word embeddings as input and applies 1-D convolutional operation on the input to extract semantic features. The features are then passed through a fully-connected layer before reaching the final softmax layer.
Thanks to the bilingual word embedding which maps the words from the two languages to the same feature representation space, the model trained in English can also be applied to documents in ILs. This enables us to use the same model and parameters for predicting SF types in both the English documents and IL documents. As described in our recent publication~\citep{Muis2018}, we primarily trained our model on English data and fine-tune it with IL annotations if they exist. Our English data include ReliefWeb dataset~\citep{Horwood2016} and LORELEI Representative Language Packs (LRLP) dataset~(\S\ref{sec:lrlp}). We extended the ReliefWeb dataset with sentences found by our bootstrapped keyword system, an extension of the Keyword Model described above with an additional keyword bootstrapping step to get more keywords. Because the resulting ReliefWeb dataset was biased by keywords, we filtered out false positives in the dataset using an SVM classifier. We used active learning strategy to rapidly build an accurate SVM classifier. More precisely, we alternately performed manual annotation and SVM training, where subsequent annotations were done on sentences for which the SVM model in the previous iteration gave low confidence score (i.e., closer to the decision boundary). This resulted in 2,562 annotations over 11 SF types. Finally, we took the top-25\% positive and negative predictions\footnote{That is, we removed 50\% of the data in the middle.} on the whole ReliefWeb dataset as our final training data.

During testing, we ran our model to each sentence in the test set to get the probability estimate of each SF type for each sentence. We then adopted two approaches to filter out SF types with low probability estimates. In the first approach, we calculated the average probability $\mu$ and standard deviation $\sigma$ for each SF type from all sentences, and filter out the predictions whose estimates fall below $\mu + \lambda\sigma$, where $\lambda$ is a hyperparameter. In our submissions we used $\lambda = -1.5$ based on the results in previous evaluations. The second approach considers the assumption that one document is not likely to describes many topics. We took only the top-$k$ SF types per document, where $k = \min(3, S)$, where $S$ is the number of sentences in the document. Upon this top-$k$ extraction, we also filter out the predictions with low probability estimates by the first approach, where we used $\lambda = 0$.

In checkpoint~2, we also experimented with the method of moment matching~\citep{Zellinger2017} to alleviate the domain mismatch between the our English training data and IL test data, aiming to build a feature extractor that only captures the semantics of the event types but not the difference in language usage between English and ILs. In other words, we tried to make the features captured by CNN informative for the event type classification and language-invariant at the same time. The method of moment matching by \citet{Zellinger2017} does this by minimizing the distance between feature vectors of English and IL text obtained from CNN. Concretely, we consider the sets of the extracted feature vectors as probability distributions and put a constraint on CNN minimizing the difference of higher order central moments of the distributions.


In checkpoint 2 we also show our Keyword Model outputs on Set1 to the NIs to be annotated with SF types, to be used as development set to estimate the performance of the numerous variants of the Neural Model. Although not perfect, since part of the training data of the Neural Model comes from a variant of the Keyword Model, we were able to identify certain hyperparameter combinations which are not performing well, and thus helping us deciding which systems to be submitted.

Like our Keyword Model, we also made another set of submissions where the IL texts were converted into IPA. For Neural Model, this also required us to train the bilingual word embeddings in IPA, after which the same SF pipeline can be applied.



\subsubsection{SF Justification}
Our two models make predictions at sentence-level, so we can simply take the sentence that was used to predict that particular SF type as our justification sentence.

\subsubsection{SF Location linking}\label{sf-location}
After populating SFs in a document, for each SF we assign the locations based on the GPEs and LOCs entities found in the sentences surrounding the justification sentence, up to $n$ sentences away, or if there is no location entities in those surrounding sentences, we assign the most recently assigned location mention. In our submissions we used $n=1$ and $n=\infty$.

\subsubsection{SF Status and Urgency detection}
In our SF systems we always predict ``insufficient'' and ``current'' for the resolution and status field, respectively, and focus our effort on urgency prediction.

For our Checkpoint 1 submissions, we had an IL-English parallel data where the English documents were labeled with sentiment, emotion and urgency labels using sentiment and pre-trained emotion systems. The sentiment system was developed using a bidirectional LSTM that was trained on English Twitter data with bilingual word embeddings. Emotion system was a gated Recurrent Neural Net trained with a multi-genre English corpus (genres: emotional blog posts, tweets, news title, movie reviews). Urgency labels were derived from emotion tag distribution (e.g., anger and fear) and according to a targeted urgency distribution. These tags are then projected to the IL documents and SVM classifier was trained in IL using sentiment as a feature. For urgency classifier in English, we combined the English documents from both IL9 and IL10 parallel data and use them as training data for English urgency classifier.

For our Checkpoint 2 submissions, we used our Checkpoint 1 classifiers to predict urgency labels on a subset of Set1, selected by taking the justification sentences of our Keyword Model when run on Set1. We combined this data with the parallel data from Set0, with 20\% stratified sub-sampling to avoid having the parallel data dominate the Set1 data. We considered two variants of the additional training data: (1) all sentences in this Set1 subset, or (2) only those sentences which confidence scores from the urgency classifier pass certain threshold. After evaluating various models with a development set we created by eliciting annotations from the NIs on Set1, we re-trained the best model with the development set as additional training data.

For English urgency classifier in Checkpoint 2, we trained an SVM classifier on both IL9 and IL10 Set S, automatically labeled by our classifier in Checkpoint 1. We also added data that we collected from Figure Eight crowdsourcing platform, which are tweets about disasters.

\section*{Acknowledgment}

This project was sponsored by the Defense Advanced Research Projects Agency (DARPA) Information Innovation Office (I2O), program: Low Resource Languages for Emergent Incidents (LORELEI), issued by DARPA/I2O under Contract No. HR0011-15-C-0114.

\bibliographystyle{bibtex/spbasic}
\bibliography{bibtex/arielsf.bib,bibtex/arielmt.bib,bibtex/references20170825.bib,bibtex/low-resource-ie.bib}

\begin{table*}[t]
\centering
\caption{CMU IL9 EDL Submissions}
\label{tab:edl_subs_il9}
\begin{tabular}{lllp{13cm}}
\toprule
System             & Condition     & CP  & Submission                                                                                                                            \\
\midrule
il9\_edl\_system01 & unconstrained & CP1 & \seqsplit{il9\_cp1\_EDL\_IL9\_sep\_bc\_if\_train\_NI\_dev\_wiki\_d8608e\_darpa\_output.conll\_Eng.tgz}                                           \\
il9\_edl\_system01 & unconstrained & CP2 & \seqsplit{il9\_cp2\_unconstrained\_EDL\_post\_updatedcp2\_IL9\_sep\_bc\_id\_cap\_961ddf\_darpa\_output.conll.tgz}                                \\
il9\_edl\_system02 & constrained   & CP1 & \seqsplit{il9\_cp1\_il9\_cp1\_EDL\_IL9\_cat\_bc\_if\_swa\_7b2a30\_darpa\_output.conll\_Eng\_Nominal.tgz}                                         \\
il9\_edl\_system02 & constrained   & CP2 & \seqsplit{il9\_cp2\_ENG\_NIL\_FINAL\_EDL\_post\_cp2\_1\_IL9\_v2\_sep\_bc\_id\_685463\_darpa\_output.conll.tgz}                                   \\
il9\_edl\_system03 & constrained   & CP1 & \seqsplit{il9\_cp1\_il9\_cp1\_EDL\_IL9\_sep\_bc\_if\_train\_NI\_dev\_wiki\_d8608e\_darpa\_output.conll\_Eng\_Nominal.tgz}                        \\
il9\_edl\_system03 & constrained   & CP2 & \seqsplit{il9\_cp2\_ENG\_NIL\_FINAL\_EDL\_post\_kb\_cp2\_IL9\_v2\_cat\_bc\_id\_a87a6c\_darpa\_output.conll.tgz}                                  \\
il9\_edl\_system04 & unconstrained & CP1 & \seqsplit{il9\_cp1\_EDL\_post\_IL9\_sep\_bc\_if\_423346\_darpa\_output.conll\_Eng.tgz}                                                           \\
il9\_edl\_system04 & unconstrained & CP2 & \seqsplit{il9\_cp2\_unconstrained\_EDL\_post\_cp2\_IL9\_sep\_bc\_id\_cap\_961ddf\_darpa\_output.conll.tgz}                                       \\
il9\_edl\_system05 & constrained   & CP1 & \seqsplit{il9\_cp1\_il9\_cp1\_EDL\_post\_IL9\_4\_sep\_bc\_if\_12f686\_darpa\_output.conll\_Eng\_Nominal.tgz}                                     \\
il9\_edl\_system05 & constrained   & CP2 & \seqsplit{il9\_cp2\_ENG\_NIL\_FINAL\_EDL\_post\_nonneg\_post\_kb\_cp2\_IL9\_v2\_cat\_bc\_id\_a87a6c\_darpa\_output.conll.tgz}                    \\
il9\_edl\_system06 & unconstrained & CP1 & \seqsplit{il9\_cp1\_EDL\_IL9\_cat\_bc\_if\_swa\_7b2a30\_darpa\_output.conll\_Eng.tgz}                                                            \\
il9\_edl\_system06 & unconstrained & CP2 & \seqsplit{il9\_cp2\_unconstrained\_EDL\_post\_updatedsetE\_IL9\_sep\_bc\_if\_2d92a1\_darpa\_output.conll.tgz}                                    \\
il9\_edl\_system07 & constrained   & CP1 & \seqsplit{il9\_cp1\_il9\_cp1\_EDL\_post\_IL9\_ensemble\_sep\_bc\_if\_2b2dc4\_darpa\_output.conll\_Eng\_Nominal.tgz}                              \\
il9\_edl\_system07 & constrained   & CP2 & \seqsplit{il9\_cp2\_ENG\_NIL\_FINAL\_EDL\_post\_nonneg\_post\_updatedsetE\_IL9\_sep\_bc\_if\_2d92a1\_darpa\_output.conll.tgz}                    \\
il9\_edl\_system08 & unconstrained & CP1 & \seqsplit{il9\_cp1\_EDL\_post\_IL9\_cat\_bc\_if\_hf\_fa5929\_darpa\_output.conll\_Eng.tgz}                                                       \\
il9\_edl\_system08 & unconstrained & CP2 & \seqsplit{il9\_cp2\_unconstrained\_EDL\_post\_cp2\_IL9\_v2\_sep\_bc\_id\_b6fcf2\_darpa\_output.conll.tgz}                                        \\
il9\_edl\_system09 & constrained   & CP1 & \seqsplit{il9\_cp1\_il9\_cp1\_EDL\_post\_IL9\_5\_sep\_bc\_if\_83cf68\_darpa\_output.conll\_Eng\_Nominal.tgz}                                     \\
il9\_edl\_system09 & constrained   & CP2 & \seqsplit{il9\_cp2\_ENG\_NIL\_FINAL\_EDL\_post\_nonneg\_post\_kb\_cp2\_IL9\_v2\_sep\_bc\_id\_b6fcf2\_darpa\_output.conll.tgz}                    \\
il9\_edl\_system10 & unconstrained & CP1 & \seqsplit{il9\_cp1\_EDL\_post\_IL9\_ensemble\_sep\_bc\_if\_2b2dc4\_darpa\_output.conll\_Eng.tgz}                                                 \\
il9\_edl\_system10 & unconstrained & CP2 & \seqsplit{il9\_cp2\_unconstrained\_EDL\_post\_cp2\_IL9\_sep\_bc\_id\_partial\_cap\_2a349f\_darpa\_output.conll.tgz}                              \\
il9\_edl\_system11 & constrained   & CP1 & \seqsplit{il9\_cp1\_il9\_cp1\_EDL\_post\_IL9\_6\_sep\_bc\_if\_511907\_darpa\_output.conll\_Eng\_Nominal.tgz}                                     \\
il9\_edl\_system11 & constrained   & CP2 & \seqsplit{il9\_cp2\_ENG\_NIL\_FINAL\_EDL\_post\_nonneg\_post\_updatedcp2\_IL9\_sep\_bc\_id\_cap\_961ddf\_darpa\_output.conll.tgz}                \\
il9\_edl\_system12 & constrained   & CP1 & \seqsplit{il9\_cp1\_il9\_cp1\_EDL\_post\_IL9\_cat\_bc\_if\_hf\_fa5929\_darpa\_output.conll\_Eng\_Nominal.tgz}                                    \\
il9\_edl\_system12 & constrained   & CP2 & \seqsplit{il9\_cp2\_ENG\_NIL\_FINAL\_EDL\_post\_nonneg\_post\_updatedcp2\_IL9\_sep\_bc\_id\_partial\_cap\_2a349f\_darpa\_output.conll.tgz}       \\
il9\_edl\_system13 & constrained   & CP1 & \seqsplit{il9\_cp1\_il9\_cp1\_EDL\_IL9\_cat\_bc\_if\_hf\_fa5929\_darpa\_output.conll\_Eng\_Nominal.tgz}                                          \\
il9\_edl\_system13 & constrained   & CP2 & \seqsplit{il9\_cp2\_ENG\_NIL\_FINAL\_EDL\_post\_cp2\_1\_IL9\_v2\_sep\_bc\_id\_004563\_darpa\_output.conll.tgz}                                   \\
il9\_edl\_system14 & unconstrained & CP1 & \seqsplit{il9\_cp1\_EDL\_IL9\_cat\_bc\_if\_hf\_fa5929\_darpa\_output.conll\_Eng.tgz}                                                             \\
il9\_edl\_system14 & unconstrained & CP2 & \seqsplit{il9\_cp2\_unconstrained\_FINAL\_EDL\_post\_cp2\_IL9\_v2\_sep\_bc\_8cde74\_darpa\_output.conll.tgz}                                     \\
il9\_edl\_system15 & unconstrained & CP1 & \seqsplit{il9\_cp1\_EDL\_post\_IL9\_6\_sep\_bc\_if\_511907\_darpa\_output.conll\_Eng.tgz}                                                        \\
il9\_edl\_system15 & unconstrained & CP2 & \seqsplit{il9\_cp2\_unconstrained\_EDL\_post\_kb\_cp2\_IL9\_v2\_sep\_bc\_id\_b6fcf2\_darpa\_output.conll.tgz}                                    \\
il9\_edl\_system16 & unconstrained & CP1 & \seqsplit{il9\_cp1\_EDL\_post\_IL9\_4\_sep\_bc\_if\_12f686\_darpa\_output.conll\_Eng.tgz}                                                        \\
il9\_edl\_system16 & unconstrained & CP2 & \seqsplit{il9\_cp2\_unconstrained\_EDL\_post\_updatedcp2\_IL9\_sep\_bc\_id\_b2f70c\_darpa\_output.conll.tgz}                                     \\
il9\_edl\_system17 & unconstrained & CP1 & \seqsplit{il9\_cp1\_EDL\_post\_IL9\_cat\_bc\_if\_swa\_7b2a30\_darpa\_output.conll\_Eng.tgz}                                                      \\
il9\_edl\_system17 & unconstrained & CP2 & \seqsplit{il9\_cp2\_unconstrained\_EDL\_post\_cp2\_IL9\_v2\_cat\_bc\_id\_a87a6c\_darpa\_output.conll.tgz}                                        \\
il9\_edl\_system18 & constrained   & CP1 & \seqsplit{il9\_cp1\_il9\_cp1\_EDL\_post\_IL9\_sep\_bc\_if\_423346\_darpa\_output.conll\_Eng\_Nominal.tgz}                                        \\
il9\_edl\_system18 & constrained   & CP2 & \seqsplit{il9\_cp2\_ENG\_NIL\_FINAL\_EDL\_post\_updated\_KB\_cp2\_IL9\_sep\_bc\_id\_partial\_cap\_2a349f\_darpa\_output.conll.tgz}               \\
il9\_edl\_system19 & constrained   & CP1 & \seqsplit{il9\_cp1\_il9\_cp1\_EDL\_post\_IL9\_cat\_bc\_if\_swa\_7b2a30\_darpa\_output.conll\_Eng\_Nominal.tgz}                                   \\
il9\_edl\_system19 & constrained   & CP2 & \seqsplit{il9\_cp2\_ENG\_NIL\_FINAL\_EDL\_post\_nonneg\_post\_updated\_KB\_cp2\_IL9\_sep\_bc\_id\_partial\_cap\_2a349f\_darpa\_output.conll.tgz} \\
il9\_edl\_system20 & unconstrained & CP1 & \seqsplit{il9\_cp1\_EDL\_post\_IL9\_5\_sep\_bc\_if\_83cf68\_darpa\_output.conll\_Eng.tgz}                                                        \\
il9\_edl\_system20 & unconstrained & CP2 & \seqsplit{il9\_cp2\_unconstrained\_EDL\_post\_nonneg\_post\_updatedcp2\_IL9\_sep\_bc\_id\_b2f70c\_darpa\_output.conll.tgz}     \\
\bottomrule                 
\end{tabular}
\end{table*}

\begin{table*}[t]
\centering
\caption{CMU IL10 EDL Submissions}
\label{tab:edl_subs_il10}
\begin{tabular}{lllp{13cm}}
\toprule
System              & Condition     & CP  & Submission                                                                                                   \\
\midrule
il10\_edl\_system01 & constrained   & CP1 & \seqsplit{il10\_cp1\_EDL2\_post\_IL10\_new\_sep\_bc\_id\_hf\_f056cb\_darpa\_output.conll\_Eng.tgz}                      \\
il10\_edl\_system01 & constrained   & CP2 & \seqsplit{il10\_cp2\_ENG\_FINAL\_EDL\_post\_cp2\_3\_IL10\_v2\_sep\_bc\_id\_81e476\_darpa\_output.conll.tgz}             \\
il10\_edl\_system02 & unconstrained & CP1 & \seqsplit{il10\_cp1\_EDL2\_IL10\_cat\_bc\_26ad4a\_darpa\_output.conll\_Eng.tgz}                                         \\
il10\_edl\_system03 & constrained   & CP1 & \seqsplit{il10\_cp1\_EDL2\_post\_IL10\_new\_cat\_bc\_id\_hf\_c8b36c\_darpa\_output.conll\_Eng.tgz}                      \\
il10\_edl\_system03 & constrained   & CP2 & \seqsplit{il10\_cp2\_ENG\_FINAL\_EDL\_post\_cp2\_1\_IL10\_v2\_sep\_bc\_id\_482e14\_darpa\_output.conll.tgz}             \\
il10\_edl\_system04 & constrained   & CP1 & \seqsplit{il10\_cp1\_EDL2\_post\_IL10\_ens\_sep\_bc\_id\_hf\_501095\_darpa\_output.conll\_Eng.tgz}                      \\
il10\_edl\_system04 & constrained   & CP2 & \seqsplit{il10\_cp2\_ENG\_FINAL\_EDL\_post\_cp2\_2\_IL10\_v2\_sep\_bc\_id\_0c0ca2\_darpa\_output.conll.tgz}             \\
il10\_edl\_system05 & unconstrained & CP1 & \seqsplit{il10\_cp1\_EDL2\_nominal.tgz}                                                                                 \\
il10\_edl\_system05 & unconstrained & CP2 & \seqsplit{il10\_cp2\_ENG\_FINAL\_EDL\_sin\_edit.conll.tgz}                                                              \\
il10\_edl\_system06 & constrained   & CP1 & \seqsplit{il10\_cp1\_EDL2\_IL10\_sep\_bc\_id\_hf\_ed334f\_darpa\_output.conll\_Eng.tgz}                                 \\
il10\_edl\_system06 & constrained   & CP2 & \seqsplit{il10\_cp2\_ENG\_FINAL\_EDL\_post\_cp2\_IL10\_v2-noipa\_sep\_bc\_id\_1c051b\_darpa\_output.conll.tgz}          \\
il10\_edl\_system07 & unconstrained & CP1 & \seqsplit{il10\_cp1\_EDL2\_IL10\_new\_cat\_bc\_id\_hf\_lm\_6a8118\_darpa\_output.conll\_Eng.tgz}                        \\
il10\_edl\_system08 & constrained   & CP1 & \seqsplit{il10\_cp1\_EDL2\_post\_IL10\_new\_cat\_bc\_if\_74855b\_darpa\_output.conll\_Eng.tgz}                          \\
il10\_edl\_system08 & constrained   & CP2 & \seqsplit{il10\_cp2\_ENG\_EDL\_post\_cp2\_IL10\_v2\_cat\_bc\_id\_faedd4\_darpa\_output.conll.tgz}                       \\
il10\_edl\_system09 & unconstrained & CP1 & \seqsplit{il10\_cp1\_EDL2\_post\_IL10\_cat\_bc\_id\_hin\_9b5243\_darpa\_output.conll\_Eng.tgz}                          \\
il10\_edl\_system09 & unconstrained & CP2 & \seqsplit{il10\_cp2\_ENG\_EDL\_post\_cp2\_IL10\_v2\_cat\_bc\_3c499e\_darpa\_output.conll.tgz}                           \\
il10\_edl\_system10 & unconstrained & CP1 & \seqsplit{il10\_cp1\_EDL2\_post\_IL10\_cat\_bc\_26ad4a\_darpa\_output.conll\_Eng.tgz}                                   \\
il10\_edl\_system11 & constrained   & CP1 & \seqsplit{il10\_cp1\_EDL2\_IL10\_new\_cat\_bc\_id\_hf\_c8b36c\_darpa\_output.conll\_Eng.tgz}                            \\
il10\_edl\_system11 & constrained   & CP2 & \seqsplit{il10\_cp2\_ENG\_FINAL\_EDL\_post\_cp2\_IL10\_v2\_sep\_bc\_id\_e244e9\_darpa\_output.conll.tgz}                \\
il10\_edl\_system12 & constrained   & CP1 & \seqsplit{il10\_cp1\_EDL2\_IL10\_new\_sep\_bc\_id\_hf\_f056cb\_darpa\_output.conll\_Eng.tgz}                            \\
il10\_edl\_system12 & constrained   & CP2 & \seqsplit{il10\_cp2\_ENG\_FINAL\_EDL\_sin\_edit3.conll.tgz}                                                             \\
il10\_edl\_system13 & unconstrained & CP1 & \seqsplit{il10\_cp1\_EDL2\_IL10\_new\_cat\_bc\_48bf80\_darpa\_output.conll\_Eng.tgz}                                    \\
il10\_edl\_system14 & unconstrained & CP1 & \seqsplit{il10\_cp1\_EDL2\_post\_IL10\_cat\_bc\_id\_hf\_4c0bcc\_darpa\_output.conll\_Eng.tgz}                           \\
il10\_edl\_system14 & unconstrained & CP2 & \seqsplit{il10\_cp2\_unconstrained\_il10\_cp2\_ENG\_EDL\_post\_cp2\_IL10\_v2\_cat\_bc\_3c499e\_darpa\_output.conll.tgz}\\
il10\_edl\_system15 & constrained   & CP1 & \seqsplit{il10\_cp1\_EDL2\_post\_IL10\_new\_cat\_bc\_id\_hf\_lm\_6a8118\_darpa\_output.conll\_Eng.tgz}                  \\
il10\_edl\_system15 & constrained   & CP2 & \seqsplit{il10\_cp2\_ENG\_FINAL\_EDL\_post\_cp2\_1\_IL10\_v2\_sep\_bc\_id\_19e5bb\_darpa\_output.conll.tgz}             \\
il10\_edl\_system17 & unconstrained & CP1 & \seqsplit{il10\_cp1\_EDL2\_post\_IL10\_cat\_bc\_id\_cce446\_darpa\_output.conll\_Eng.tgz}                               \\
il10\_edl\_system18 & constrained   & CP1 & \seqsplit{il10\_cp1\_EDL2\_post\_IL10\_sep\_bc\_id\_hf\_ed334f\_darpa\_output.conll\_Eng.tgz}                           \\
il10\_edl\_system18 & constrained   & CP2 & \seqsplit{il10\_cp2\_ENG\_EDL\_post\_cp2\_IL10\_sep\_bc\_id\_58c8ed\_darpa\_output.conll.tgz}                           \\
il10\_edl\_system19 & constrained   & CP1 & \seqsplit{il10\_cp1\_EDL2\_IL10\_new\_cat\_bc\_if\_74855b\_darpa\_output.conll\_Eng.tgz}                                \\
il10\_edl\_system19 & constrained   & CP2 & \seqsplit{il10\_cp2\_ENG\_FINAL\_EDL\_post\_cp2\_IL10\_v2\_sep\_bc\_d665c8\_darpa\_output.conll.tgz}                    \\
il10\_edl\_system20 & unconstrained & CP1 & \seqsplit{il10\_cp1\_EDL2\_IL10\_cat\_bc\_id\_cce446\_darpa\_output.conll\_Eng.tgz} \\
\bottomrule                                   
\end{tabular}
\end{table*}

\begin{table*}[t]
\centering
\caption{CMU IL9 MT Submissions}
\label{tab:mt_subs_il9}
\begin{tabular}{lllp{13cm}}
\toprule
System            & Condition     & CP  & Submission                                                                     \\
\midrule
il9\_mt\_system1  & constrained   & CP1 & \seqsplit{cdec-misalign-v1\_il9\_cp1.tgz}                                                \\
il9\_mt\_system1  & constrained   & CP2 & \seqsplit{cdec-lemma-punk-v4\_il9\_cp2.tgz}                                               \\
il9\_mt\_system2  & constrained   & CP1 & \seqsplit{cdec-align-nobrown-v2\_il9\_cp1.tgz}                                            \\
il9\_mt\_system2  & constrained   & CP2 & \seqsplit{cdec-lemmav7-clitics\_il9\_cp2.tgz}                                             \\
il9\_mt\_system3  & constrained   & CP1 & \seqsplit{s2s-tok.final-v3\_il9\_cp1.tgz}                                                 \\
il9\_mt\_system3  & constrained   & CP2 & \seqsplit{cdec-new5-c600-tok\_il9\_cp2.tgz}                                               \\
il9\_mt\_system4  & constrained   & CP1 & \seqsplit{s2s-spm8k.final-v4\_il9\_cp1.tgz}                                               \\
il9\_mt\_system4  & constrained   & CP2 & \seqsplit{cdec-tok-punk-v4\_il9\_cp2.tgz}                                                 \\
il9\_mt\_system5  & constrained   & CP1 & \seqsplit{s2s-spm16k.final-v5\_il9\_cp1.tgz}                                              \\
il9\_mt\_system5  & constrained   & CP2 & \seqsplit{kineng-ensemble1.eval.setE\_dednt\_il9\_cp2.tgz}                                \\
il9\_mt\_system6  & unconstrained & CP2 & \seqsplit{unconstrained\_cdec-new4-c600-lemma\_il9\_cp2.tgz}                              \\
il9\_mt\_system7  & constrained   & CP1 & \seqsplit{nolor1.def.adapt-kinpl.eval.setE\_dednt\_il9\_cp1.tgz}                          \\
il9\_mt\_system7  & constrained   & CP2 & \seqsplit{kineng-ensemble1.regp1.2.eval.setE\_dednt\_il9\_cp2.tgz}                        \\
il9\_mt\_system8  & unconstrained & CP2 & \seqsplit{unconstrained\_cdec-new5-c200-lemma\_il9\_cp2.tgz}                              \\
il9\_mt\_system9  & unconstrained & CP2 & \seqsplit{unconstrained\_cdec-new5-c600-lemma\_il9\_cp2.tgz}                              \\
il9\_mt\_system10 & constrained   & CP1 & \seqsplit{kinp1.def.eval.setE\_dednt\_il9\_cp1.tgz}                                       \\
il9\_mt\_system10 & constrained   & CP2 & \seqsplit{kineng-ensemble2.eval.setE\_dednt\_il9\_cp2.tgz}                                \\
il9\_mt\_system11 & unconstrained & CP2 & \seqsplit{unconstrained\_cdec-tok-new5-c200\_il9\_cp2.tgz}                                \\
il9\_mt\_system12 & unconstrained & CP2 & \seqsplit{unconstrained\_kinengpl2ne11.def-jointseg.eval.setE\_dednt\_il9\_cp2.tgz}       \\
il9\_mt\_system13 & unconstrained & CP2 & \seqsplit{unconstrained\_kinengpl21.def.avg.new\_dnt.eval.setE\_dednt\_il9\_cp2.tgz}      \\
il9\_mt\_system14 & constrained   & CP1 & \seqsplit{cdec-realign-nobrown-v6\_il9\_cp1.tgz}                                          \\
il9\_mt\_system14 & constrained   & CP2 & \seqsplit{kineng-ensemble3.eval.setE\_dednt\_il9\_cp2.tgz}                                \\
il9\_mt\_system15 & constrained   & CP1 & \seqsplit{all.ipajoint.decode.b5.setE\_dednt\_il9\_cp1.tgz}                               \\
il9\_mt\_system15 & constrained   & CP2 & \seqsplit{kinfraeng-ensemble1.eval.setE\_dednt\_il9\_cp2.tgz}                             \\
il9\_mt\_system16 & constrained   & CP1 & \seqsplit{cdec-realign-v7\_il9\_cp1.tgz}                                                 \\
il9\_mt\_system16 & constrained   & CP2 & \seqsplit{memt.kin.setE\_dednt\_il9\_cp2.tgz}                                             \\
il9\_mt\_system17 & unconstrained & CP2 & \seqsplit{unconstrained\_nolor1.def.adapt-kinengpl2.eval.setE\_dednt\_il9\_cp2.tgz}       \\
il9\_mt\_system18 & unconstrained & CP2 & \seqsplit{unconstrained\_nolor1.def.adapt-kinengpl2ne1.eval.setE\_dednt\_il9\_cp2.tgz}    \\
il9\_mt\_system19 & unconstrained & CP2 & \seqsplit{unconstrained\_all-aug.ipajoint.decode.b5.setE\_dednt\_il9\_cp2.tgz}            \\
il9\_mt\_system20 & unconstrained & CP2 & \seqsplit{unconstrained\_nolor1.def.adapt-kinfraengpl2ne1.eval.setE\_dednt\_il9\_cp2.tgz}\\
\bottomrule
\end{tabular}
\end{table*}

\begin{table*}[t]
\centering
\caption{CMU IL10 MT Submissions}
\label{tab:mt_subs_il10}
\begin{tabular}{lllp{13cm}}
\toprule
System             & Condition     & CP  & Submission                                                                 \\
\midrule
il10\_mt\_system01 & constrained   & CP1 & \seqsplit{cdec-misalign-v1\_il10\_cp1.tgz}                                            \\
il10\_mt\_system01 & constrained   & CP2 & \seqsplit{cdec-lemma-splitsv3\_il10\_cp2.tgz}                                         \\
il10\_mt\_system02 & constrained   & CP2 & \seqsplit{sin-ensemble1.regy1.2.newdnt.eval.setE\_dednt\_il10\_cp2.tgz}               \\
il10\_mt\_system03 & constrained   & CP2 & \seqsplit{sinhinben-ensemble1.eval.setE\_dednt\_il10\_cp2.tgz}                        \\
il10\_mt\_system04 & constrained   & CP1 & \seqsplit{nolor1.def.adapt-sinp.eval.setE\_dednt\_il10\_cp1.tgz}                      \\
il10\_mt\_system04 & constrained   & CP2 & \seqsplit{cdec-tok-splitsv3\_il10\_cp2.tgz}                                           \\
il10\_mt\_system05 & unconstrained & CP2 & \seqsplit{unconstrained\_all-T.ipajoint.decode.b5.setE\_dednt\_il10\_cp2.tgz}         \\
il10\_mt\_system06 & constrained   & CP1 & \seqsplit{sinp1.def.eval.setE\_dednt\_il10\_cp1.tgz}                                  \\
il10\_mt\_system06 & constrained   & CP2 & \seqsplit{cdec-tok-splitsv3-spm\_il10\_cp2.tgz}                                       \\
il10\_mt\_system07 & constrained   & CP2 & \seqsplit{memt.sin.setE\_dednt\_il10\_cp2.tgz}                                        \\
il10\_mt\_system08 & unconstrained & CP2 & \seqsplit{unconstrained\_cdec-tok-splitsv2\_il10\_cp2.tgz}                            \\
il10\_mt\_system09 & unconstrained & CP2 & \seqsplit{unconstrained\_cdec-tok-splitsv3-v2\_il10\_cp2.tgz}                         \\
il10\_mt\_system10 & constrained   & CP2 & \seqsplit{nolor1.def.adapt-sinp.eval.setE\_dednt\_il10\_cp2.tgz}                      \\
il10\_mt\_system11 & unconstrained & CP2 & \seqsplit{unconstrained\_nolor1.def.adapt-sinhinbenp.eval.setE\_dednt\_il10\_cp2.tgz} \\
il10\_mt\_system12 & constrained   & CP2 & \seqsplit{nolor1.def.adapt-sinpl.eval.setE\_dednt\_il10\_cp2.tgz}                     \\
il10\_mt\_system13 & unconstrained & CP2 & \seqsplit{unconstrained\_nolor1.def-r2.adapt-sinpl.eval.setE\_dednt\_il10\_cp2.tgz}   \\
il10\_mt\_system14 & unconstrained & CP2 & \seqsplit{unconstrained\_sin-ensemble1.eval.setE\_dednt\_il10\_cp2.tgz}               \\
il10\_mt\_system15 & unconstrained & CP2 & \seqsplit{unconstrained\_sin-ensemble1.new\_dnt.eval.setE\_dednt\_il10\_cp2.tgz}      \\
il10\_mt\_system16 & unconstrained & CP2 & \seqsplit{unconstrained\_sinhinbenp1.def.eval.setE\_dednt\_il10\_cp2.tgz}             \\
il10\_mt\_system17 & constrained   & CP2 & \seqsplit{sin-ensemble1.eval.setE\_dednt\_il10\_cp2.tgz}                              \\
il10\_mt\_system18 & unconstrained & CP2 & \seqsplit{unconstrained\_sinp1.def.eval.setE\_dednt\_il10\_cp2.tgz}                   \\
il10\_mt\_system19 & constrained   & CP2 & \seqsplit{sin-ensemble1.regy1.2.eval.setE\_dednt\_il10\_cp2.tgz}                      \\
il10\_mt\_system20 & unconstrained & CP2 & \seqsplit{unconstrained\_sinpl1.def.eval.setE\_dednt\_il10\_cp2.tgz}                 \\
\bottomrule
\end{tabular}
\end{table*}

\begin{table*}[t]
\centering
\caption{CMU IL9 SF Submissions}
\label{tab:sf_subs_il9}
\begin{tabular}{lllp{13cm}}
\toprule
System            & Condition     & CP  & Submission    \\
\midrule
il9\_sf\_system01 & unconstrained & CP2 & \seqsplit{unconstrained\_edl\_output\_v6\_outputs\_il9\_il9\_all\_uniqlast\_place\_sys\_ner\_locforce\_remove\_status\_loc\_final\_cbia\_urg-self-v4-setS\_IL9\_NN\_CP2\_submissions\_v1.noNI.ortho.dom\_simple.v1.thres.-1.5\_il9\_cp2.tgz}                      \\
il9\_sf\_system02 & unconstrained & CP2 & \seqsplit{unconstrained\_edl\_output\_v6\_outputs\_il9\_il9\_sys\_nb1\_na1\_place\_sys\_ner\_locforce\_remove\_status\_loc\_final\_cbia\_urg-self-v4-setS\_IL9\_KWD\_CP2\_submissions\_IL9\_kin\_setE\_NI\_MT\_ipav2\_ignorev2\_speech\_th\_0.8\_il9\_cp2.tgz}    \\
il9\_sf\_system03 & constrained   & CP1 & \seqsplit{edl\_output\_v2\_outputs\_il9\_il9\_sys\_nb1\_na1\_place\_sys\_ner\_status\_loc\_code\_kw\_outputs\_il9\_kin\_setE\_il9\_kin\_setE\_sfs\_KWD1\_k\_2\_th\_0.9\_il9\_cp1.tgz}                                                                             \\
il9\_sf\_system03 & constrained   & CP2 & \seqsplit{edl\_output\_v6\_outputs\_il9\_il9\_all\_uniqlast\_place\_sys\_ner\_locforce\_remove\_status\_loc\_final\_cbia\_urg-self-v4-setS\_IL9\_KWD\_CP2\_submissions\_IL9\_kin\_setE\_NI\_MT\_orthov2\_ignorev2\_speech\_th\_0.8\_il9\_cp2.tgz}                 \\
il9\_sf\_system04 & unconstrained & CP2 & \seqsplit{unconstrained\_edl\_output\_v6\_outputs\_il9\_il9\_sys\_nb1\_na1\_place\_sys\_ner\_locforce\_remove\_status\_loc\_final\_cbia\_urg-self-v4-setS\_IL9\_KWD\_CP2\_submissions\_IL9\_kin\_setE\_NI\_only\_ipav2\_speech\_th\_0.8\_il9\_cp2.tgz}            \\
il9\_sf\_system05 & constrained   & CP1 & \seqsplit{edl\_output\_v2\_outputs\_il9\_il9\_sys\_nb1\_na1\_place\_sys\_ner\_status\_loc\_code\_kw\_outputs\_il9\_kin\_setE\_il9\_kin\_setE\_sfs\_KWD1\_k\_2\_th\_0.8\_il9\_cp1.tgz}                                                                             \\
il9\_sf\_system05 & constrained   & CP2 & \seqsplit{edl\_output\_v6\_outputs\_il9\_il9\_all\_uniqlast\_place\_sys\_ner\_locforce\_remove\_status\_loc\_final\_cbia\_urg-self-v4-setS\_IL9\_NN\_CP2\_submissions\_v1.NI.ortho.dom\_simple.postproc.v1.aggmax.meanthresh.top1-3\_il9\_cp2.tgz}                \\
il9\_sf\_system06 & unconstrained & CP2 & \seqsplit{unconstrained\_edl\_output\_v6\_outputs\_il9\_il9\_sys\_nb1\_na1\_place\_sys\_ner\_locforce\_remove\_status\_loc\_final\_cbia\_urg-self-v4-setS\_IL9\_KWD\_CP2\_submissions\_IL9\_kin\_setE\_NI\_only\_kinmorphv7\_speech\_th\_0.8\_il9\_cp2.tgz}       \\
il9\_sf\_system07 & constrained   & CP1 & \seqsplit{edl\_output\_v2\_outputs\_il9\_il9\_sys\_nb0\_na0\_place\_sys\_ner\_status\_loc\_cbia\_urg-all\_code\_kw\_outputs\_il9\_kin\_setE\_il9\_kin\_setE\_sfs\_KWD1\_k\_2\_th\_0.9\_il9\_cp1.tgz}                                                              \\
il9\_sf\_system07 & constrained   & CP2 & \seqsplit{edl\_output\_v6\_outputs\_il9\_il9\_sys\_nb1\_na1\_place\_sys\_ner\_locforce\_remove\_status\_loc\_final\_cbia\_urg-self-v4-setS\_IL9\_KWD\_CP2\_submissions\_IL9\_kin\_setE\_NI\_MT\_ipav2\_ignorev2\_speech\_th\_0.9\_il9\_cp2.tgz}                   \\
il9\_sf\_system08 & unconstrained & CP2 & \seqsplit{unconstrained\_edl\_output\_v6\_outputs\_il9\_il9\_sys\_nb1\_na1\_place\_sys\_ner\_locforce\_remove\_status\_loc\_final\_cbia\_urg-self-v4-setS\_IL9\_NN\_CP2\_submissions\_v1.NI.ipa.dom\_simple.v1.thres.-1.5\_il9\_cp2.tgz}                          \\
il9\_sf\_system09 & constrained   & CP1 & \seqsplit{edl\_output\_v2\_outputs\_il9\_il9\_sys\_nb0\_na0\_place\_sys\_ner\_status\_loc\_cbia\_urg-all\_code\_kw\_outputs\_il9\_kin\_setE\_il9\_kin\_setE\_sfs\_KWD1\_k\_2\_th\_0.8\_il9\_cp1.tgz}                                                              \\
il9\_sf\_system09 & constrained   & CP2 & \seqsplit{edl\_output\_v6\_outputs\_il9\_il9\_sys\_nb1\_na1\_place\_sys\_ner\_locforce\_remove\_status\_loc\_final\_cbia\_urg-self-v4-setS\_IL9\_KWD\_CP2\_submissions\_IL9\_kin\_setE\_NI\_MT\_orthov2\_ignorev2\_speech\_th\_0.8\_il9\_cp2.tgz}                 \\
il9\_sf\_system10 & unconstrained & CP2 & \seqsplit{unconstrained\_edl\_output\_v6\_outputs\_il9\_il9\_sys\_nb1\_na1\_place\_sys\_ner\_locforce\_remove\_status\_loc\_final\_cbia\_urg-self-v4-setS\_IL9\_NN\_CP2\_submissions\_v1.NI.lemma.dom\_simple.postproc.v1.aggmax.meanthresh.top1-3\_il9\_cp2.tgz} \\
il9\_sf\_system11 & constrained   & CP1 & \seqsplit{edl\_output\_v2\_outputs\_il9\_il9\_all\_uniqlast\_place\_sys\_ner\_status\_loc\_cbia\_urg-all\_code\_kw\_outputs\_il9\_kin\_setE\_il9\_kin\_setE\_sfs\_KWD1\_k\_2\_th\_0.8\_il9\_cp1.tgz}                                                              \\
il9\_sf\_system11 & constrained   & CP2 & \seqsplit{edl\_output\_v6\_outputs\_il9\_il9\_sys\_nb1\_na1\_place\_sys\_ner\_locforce\_remove\_status\_loc\_final\_cbia\_urg-self-v4-setS\_IL9\_KWD\_CP2\_submissions\_IL9\_kin\_setE\_NI\_only\_orthov2\_speech\_th\_0.8\_il9\_cp2.tgz}                         \\
il9\_sf\_system12 & constrained   & CP1 & \seqsplit{edl\_output\_v2\_outputs\_il9\_il9\_all\_uniqlast\_place\_sys\_ner\_status\_loc\_cbia\_urg-all\_code\_kw\_outputs\_il9\_kin\_setE\_il9\_kin\_setE\_sfs\_KWD1\_k\_2\_th\_0.9\_il9\_cp1.tgz}                                                              \\
il9\_sf\_system12 & constrained   & CP2 & \seqsplit{edl\_output\_v6\_outputs\_il9\_il9\_sys\_nb1\_na1\_place\_sys\_ner\_locforce\_remove\_status\_loc\_final\_cbia\_urg-self-v4-setS\_IL9\_NN\_CP2\_submissions\_v1.NI.ortho.dom\_mmatch.v1.thres.0\_il9\_cp2.tgz}                                          \\
il9\_sf\_system13 & unconstrained & CP2 & \seqsplit{unconstrained\_edl\_output\_v6\_outputs\_il9\_il9\_sys\_nb1\_na1\_place\_sys\_ner\_locforce\_remove\_status\_loc\_final\_cbia\_urg-self-v4-setS\_IL9\_NN\_CP2\_submissions\_v1.NI.ortho.dom\_simple.v1.thres.-1.5\_il9\_cp2.tgz}                        \\
il9\_sf\_system14 & unconstrained & CP2 & \seqsplit{unconstrained\_edl\_output\_v6\_outputs\_il9\_il9\_sys\_nb1\_na1\_place\_sys\_ner\_locforce\_remove\_status\_loc\_final\_IL9\_KWD\_CP2\_submissions\_IL9\_kin\_setE\_NI\_MT\_orthov2\_ignorev2\_speech\_th\_0.8\_il9\_cp2.tgz}                          \\
il9\_sf\_system15 & unconstrained & CP2 & \seqsplit{unconstrained\_edl\_output\_v6\_outputs\_il9\_il9\_sys\_nb1\_na1\_place\_sys\_ner\_locforce\_remove\_status\_loc\_final\_IL9\_KWD\_CP2\_submissions\_IL9\_kin\_setE\_NI\_only\_kinmorphv7\_speech\_th\_0.8\_il9\_cp2.tgz}                               \\
il9\_sf\_system16 & unconstrained & CP2 & \seqsplit{unconstrained\_edl\_output\_v6\_outputs\_il9\_il9\_sys\_nb1\_na1\_place\_sys\_ner\_locforce\_remove\_status\_loc\_final\_IL9\_NN\_CP2\_submissions\_v1.NI.ortho.dom\_simple.postproc.v1.aggmax.meanthresh.top1-3\_il9\_cp2.tgz}                         \\
il9\_sf\_system17 & constrained   & CP2 & \seqsplit{edl\_output\_v6\_outputs\_il9\_il9\_sys\_nb1\_na1\_place\_sys\_ner\_locforce\_remove\_status\_loc\_final\_cbia\_urg-self-v4-setS\_IL9\_NN\_CP2\_submissions\_v1.NI.ortho.dom\_simple.postproc.v1.aggmax.meanthresh.top1-3\_il9\_cp2.tgz}                \\
il9\_sf\_system18 & constrained   & CP2 & \seqsplit{edl\_output\_v6\_outputs\_il9\_il9\_sys\_nb1\_na1\_place\_sys\_ner\_locforce\_remove\_status\_loc\_final\_cbia\_urg-self-v4-setS\_IL9\_NN\_CP2\_submissions\_v1.noNI.lemma.dom\_simple.v1.thres.-1.5\_il9\_cp2.tgz}                                     \\
il9\_sf\_system19 & constrained   & CP2 & \seqsplit{edl\_output\_v6\_outputs\_il9\_il9\_sys\_nb1\_na1\_place\_sys\_ner\_locforce\_remove\_status\_loc\_final\_cbia\_urg-self-v4-setS\_IL9\_NN\_CP2\_submissions\_v1.noNI.ortho.dom\_simple.v1.thres.-1.5\_il9\_cp2.tgz}                                     \\
 il9\_sf\_system20 & constrained   & CP2 & \seqsplit{edl\_output\_v6\_outputs\_il9\_il9\_sys\_nb1\_na1\_place\_sys\_ner\_locforce\_remove\_status\_loc\_final\_IL9\_NN\_CP2\_submissions\_v1.noNI.ortho.dom\_simple.v1.thres.-1.5\_il9\_cp2.tgz}                                                     \\
\bottomrule
\end{tabular}
\end{table*}

\begin{table*}[t]
\centering
\tabcolsep=4pt
\caption{CMU IL10 SF Submissions}
\label{tab:sf_subs_il10}
\begin{tabular}{lllp{13.75cm}}
\toprule
System             & Condition     & CP  & Submission  \\
\midrule
il10\_sf\_system0  & constrained   & CP1 & \seqsplit{edl\_output\_v1\_outputs\_il10\_il10\_sys\_nb1\_na1\_place\_sys\_ner\_status\_loc\_xlingual\_bwe\_XlingualEmb\_dom\_simple\_tgt\_train\_0\_KWD1\_k\_2\_th\_0.8\_out.thres.-3\_il10\_cp1.tgz}                                                                \\
il10\_sf\_system0  & constrained   & CP2 & \seqsplit{edl\_output\_v7\_outputs\_il10\_il10\_all\_uniqlast\_place\_sys\_ner\_locforce\_remove\_status\_loc\_final\_cbia\_urg-self-v4-setS\_IL10\_NN\_CP2\_submissions\_v1.NI.ortho.dom\_simple.postproc.v1.aggmax.meanthresh.top1-3\_il10\_cp2.tgz}                \\
il10\_sf\_system01 & constrained   & CP1 & \seqsplit{edl\_output\_v1\_outputs\_il10\_il10\_sys\_nb1\_na1\_place\_sys\_ner\_status\_loc\_xlingual\_bwe\_XlingualEmb\_dom\_simple\_tgt\_train\_0\_KWD1\_k\_2\_th\_0.8\_out.thres.-1.5\_il10\_cp1.tgz}                                                              \\
il10\_sf\_system01 & constrained   & CP2 & \seqsplit{edl\_output\_v7\_outputs\_il10\_il10\_sys\_nb1\_na1\_place\_sys\_ner\_locforce\_remove\_status\_loc\_final\_cbia\_urg-self-v4-setS\_IL10\_KWD\_CP2\_submissions\_IL10\_sin\_setE\_NI\_MT\_orthov3\_ignorev2\_speech\_th\_0.8\_il10\_cp2.tgz}                \\
il10\_sf\_system02 & constrained   & CP1 & \seqsplit{edl\_output\_v1\_outputs\_il10\_il10\_sys\_nb1\_na1\_place\_sys\_ner\_status\_loc\_xlingual\_bwe\_XlingualEmb\_dom\_simple\_tgt\_train\_0\_KWD1\_k\_2\_th\_0.8\_out.aggmax.gmeanthresh.top1-3\_il10\_cp1.tgz}                                               \\
il10\_sf\_system02 & constrained   & CP2 & \seqsplit{edl\_output\_v7\_outputs\_il10\_il10\_sys\_nb1\_na1\_place\_sys\_ner\_locforce\_remove\_status\_loc\_final\_cbia\_urg-self-v4-setS\_IL10\_KWD\_CP2\_submissions\_IL10\_sin\_setE\_NI\_MT\_orthov3\_ignorev2\_speech\_th\_0.9\_il10\_cp2.tgz}                \\
il10\_sf\_system03 & unconstrained & CP1 & \seqsplit{edl\_output\_v1\_outputs\_il10\_il10\_sys\_nb1\_na1\_place\_sys\_ner\_status\_loc\_code\_kw\_outputs\_il10\_sin\_setE\_il10\_sin\_setE\_sfs\_KWD1\_k\_2\_th\_0.9\_il10\_cp1.tgz}                                                                            \\
il10\_sf\_system03 & unconstrained & CP2 & \seqsplit{unconstrained\_edl\_output\_v7\_outputs\_il10\_il10\_all\_uniqlast\_place\_sys\_ner\_locforce\_remove\_status\_loc\_final\_cbia\_urg-self-v4-setS\_IL10\_KWD\_CP2\_submissions\_IL10\_sin\_setE\_NI\_only\_orthov3\_speech\_th\_0.8\_il10\_cp2.tgz}         \\
il10\_sf\_system04 & unconstrained & CP1 & \seqsplit{edl\_output\_v1\_outputs\_il10\_il10\_sys\_nb1\_na1\_place\_sys\_ner\_status\_loc\_code\_kw\_outputs\_il10\_sin\_setE\_il10\_sin\_setE\_sfs\_KWD1\_k\_2\_th\_0.8\_il10\_cp1.tgz}                                                                            \\
il10\_sf\_system04 & unconstrained & CP2 & \seqsplit{unconstrained\_edl\_output\_v7\_outputs\_il10\_il10\_all\_uniqlast\_place\_sys\_ner\_locforce\_remove\_status\_loc\_final\_cbia\_urg-self-v4-setS\_IL10\_NN\_CP2\_submissions\_v1.noNI.ortho.dom\_simple.v1.thres.-1.5\_il10\_cp2.tgz}                      \\
il10\_sf\_system05 & constrained   & CP1 & \seqsplit{edl\_output\_v1\_outputs\_il10\_il10\_sys\_nb1\_na1\_place\_sys\_ner\_status\_loc\_cbia\_urg-all\_code\_kw\_outputs\_il10\_sin\_setE\_il10\_sin\_setE\_sfs\_KWD1\_k\_2\_th\_0.9\_il10\_cp1.tgz}                                                             \\
il10\_sf\_system05 & constrained   & CP2 & \seqsplit{edl\_output\_v7\_outputs\_il10\_il10\_sys\_nb1\_na1\_place\_sys\_ner\_locforce\_remove\_status\_loc\_final\_cbia\_urg-self-v4-setS\_IL10\_NN\_CP2\_submissions\_v1.NI.ortho.dom\_mmatch.v1.thres.0\_il10\_cp2.tgz}                                          \\
il10\_sf\_system06 & constrained   & CP1 & \seqsplit{edl\_output\_v1\_outputs\_il10\_il10\_sys\_nb1\_na1\_place\_sys\_ner\_status\_loc\_cbia\_urg-all\_code\_kw\_outputs\_il10\_sin\_setE\_il10\_sin\_setE\_sfs\_KWD1\_k\_2\_th\_0.8\_il10\_cp1.tgz}                                                             \\
il10\_sf\_system06 & constrained   & CP2 & \seqsplit{edl\_output\_v7\_outputs\_il10\_il10\_sys\_nb1\_na1\_place\_sys\_ner\_locforce\_remove\_status\_loc\_final\_cbia\_urg-self-v4-setS\_IL10\_NN\_CP2\_submissions\_v1.NI.ortho.dom\_simple.postproc.v1.aggmax.meanthresh.top1-3\_il10\_cp2.tgz}                \\
il10\_sf\_system07 & unconstrained & CP1 & \seqsplit{edl\_output\_v1\_outputs\_il10\_il10\_sys\_nb0\_na0\_place\_sys\_ner\_status\_loc\_cbia\_urg-all\_code\_kw\_outputs\_il10\_sin\_setE\_il10\_sin\_setE\_sfs\_KWD1\_k\_2\_th\_0.9\_il10\_cp1.tgz}                                                             \\
il10\_sf\_system07 & unconstrained & CP2 & \seqsplit{unconstrained\_edl\_output\_v7\_outputs\_il10\_il10\_sys\_nb1\_na1\_place\_sys\_ner\_locforce\_remove\_status\_loc\_final\_cbia\_urg-self-v4-setS\_IL10\_KWD\_CP2\_submissions\_IL10\_sin\_setE\_NI\_MT\_ignorev2\_sinmorph4\_speech\_th\_0.8\_il1.tgz}     \\
il10\_sf\_system08 & constrained   & CP1 & \seqsplit{edl\_output\_v1\_outputs\_il10\_il10\_sys\_nb0\_na0\_place\_sys\_ner\_status\_loc\_cbia\_urg-all\_code\_kw\_outputs\_il10\_sin\_setE\_il10\_sin\_setE\_sfs\_KWD1\_k\_2\_th\_0.8\_il10\_cp1.tgz}                                                             \\
il10\_sf\_system08 & constrained   & CP2 & \seqsplit{edl\_output\_v7\_outputs\_il10\_il10\_sys\_nb1\_na1\_place\_sys\_ner\_locforce\_remove\_status\_loc\_final\_IL10\_KWD\_CP2\_submissions\_IL10\_sin\_setE\_NI\_MT\_orthov3\_ignorev2\_speech\_th\_0.8\_il10\_cp2.tgz}                                        \\
il10\_sf\_system09 & unconstrained & CP1 & \seqsplit{edl\_output\_v1\_outputs\_il10\_il10\_all\_uniqlast\_place\_sys\_ner\_status\_loc\_cbia\_urg-all\_code\_kw\_outputs\_il10\_sin\_setE\_il10\_sin\_setE\_sfs\_KWD1\_k\_2\_th\_0.9\_il10\_cp1.tgz}                                                             \\
il10\_sf\_system09 & unconstrained & CP2 & \seqsplit{unconstrained\_edl\_output\_v7\_outputs\_il10\_il10\_sys\_nb1\_na1\_place\_sys\_ner\_locforce\_remove\_status\_loc\_final\_cbia\_urg-self-v4-setS\_IL10\_KWD\_CP2\_submissions\_IL10\_sin\_setE\_NI\_MT\_ipav3\_ignorev2\_speech\_th\_0.8\_il10\_cp.tgz}    \\
il10\_sf\_system10 & unconstrained & CP1 & \seqsplit{edl\_output\_v1\_outputs\_il10\_il10\_all\_uniqlast\_place\_sys\_ner\_status\_loc\_cbia\_urg-all\_code\_kw\_outputs\_il10\_sin\_setE\_il10\_sin\_setE\_sfs\_KWD1\_k\_2\_th\_0.8\_il10\_cp1.tgz}                                                             \\
il10\_sf\_system10 & unconstrained & CP2 & \seqsplit{unconstrained\_edl\_output\_v7\_outputs\_il10\_il10\_all\_uniqlast\_place\_sys\_ner\_locforce\_remove\_status\_loc\_final\_cbia\_urg-self-v4-setS\_IL10\_KWD\_CP2\_submissions\_IL10\_sin\_setE\_NI\_MT\_orthov3\_ignorev2\_speech\_th\_0.8\_il10\_cp2.tgz} \\
il10\_sf\_system11 & constrained   & CP1 & \seqsplit{edl\_output\_v1\_outputs\_il10\_il10\_sys\_nb1\_na1\_place\_sys\_ner\_status\_loc\_speech\_v2\_code\_kw\_outputs\_il10\_sin\_setE\_il10\_sin\_setE\_sfs\_KWD1\_k\_2\_th\_0.8\_il10\_cp1.tgz}                                                                \\
il10\_sf\_system11 & constrained   & CP2 & \seqsplit{edl\_output\_v7\_outputs\_il10\_il10\_sys\_nb1\_na1\_place\_sys\_ner\_locforce\_remove\_status\_loc\_final\_cbia\_urg-self-v4-setS\_IL10\_NN\_CP2\_submissions\_v1.noNI.ortho.dom\_simple.v1.thres.-1.5\_il10\_cp2.tgz}                                     \\
il10\_sf\_system12 & unconstrained & CP2 & \seqsplit{unconstrained\_edl\_output\_v7\_outputs\_il10\_il10\_sys\_nb1\_na1\_place\_sys\_ner\_locforce\_remove\_status\_loc\_final\_cbia\_urg-self-v4-setS\_IL10\_NN\_CP2\_submissions\_v1.NI.lemma.dom\_simple.postproc.v1.aggmax.meanthresh.top1-3\_il10\_cp2.tgz} \\
il10\_sf\_system13 & unconstrained & CP2 & \seqsplit{unconstrained\_edl\_output\_v7\_outputs\_il10\_il10\_sys\_nb1\_na1\_place\_sys\_ner\_locforce\_remove\_status\_loc\_final\_cbia\_urg-self-v4-setS\_IL10\_NN\_CP2\_submissions\_v1.NI.ortho.dom\_simple.v1.thres.-1.5\_il10\_cp2.tgz}                        \\
il10\_sf\_system14 & constrained   & CP1 & \seqsplit{edl\_output\_v1\_outputs\_il10\_il10\_all\_uniqlast\_place\_sys\_ner\_status\_loc\_cbia\_urg-all\_speech\_v2\_code\_kw\_outputs\_il10\_sin\_setE\_il10\_sin\_setE\_sfs\_KWD1\_k\_2\_th\_0.8\_il10\_cp1.tgz}                                                 \\
il10\_sf\_system14 & constrained   & CP2 & \seqsplit{edl\_output\_v7\_outputs\_il10\_il10\_sys\_nb1\_na1\_place\_sys\_ner\_locforce\_remove\_status\_loc\_final\_cbia\_urg-self-v4-setS\_IL10\_KWD\_CP2\_submissions\_IL10\_sin\_setE\_NI\_only\_orthov3\_speech\_th\_0.8\_il10\_cp2.tgz}                        \\
il10\_sf\_system15 & unconstrained & CP2 & \seqsplit{unconstrained\_edl\_output\_v7\_outputs\_il10\_il10\_sys\_nb1\_na1\_place\_sys\_ner\_locforce\_remove\_status\_loc\_final\_IL10\_KWD\_CP2\_submissions\_IL10\_sin\_setE\_NI\_only\_orthov3\_speech\_th\_0.8\_il10\_cp2.tgz}                                 \\
il10\_sf\_system16 & constrained   & CP1 & \seqsplit{edl\_output\_v1\_outputs\_il10\_il10\_sys\_nb1\_na1\_place\_sys\_ner\_status\_loc\_speech\_v2\_code\_kw\_outputs\_il10\_sin\_setE\_il10\_sin\_setE\_sfs\_KWD1\_k\_2\_th\_0.9\_il10\_cp1.tgz}                                                                \\
il10\_sf\_system16 & constrained   & CP2 & \seqsplit{edl\_output\_v7\_outputs\_il10\_il10\_sys\_nb1\_na1\_place\_sys\_ner\_locforce\_remove\_status\_loc\_final\_IL10\_NN\_CP2\_submissions\_v1.NI.ortho.dom\_simple.postproc.v1.aggmax.meanthresh.top1-3\_il10\_cp2.tgz}                                        \\
il10\_sf\_system17 & unconstrained & CP2 & \seqsplit{unconstrained\_edl\_output\_v7\_outputs\_il10\_il10\_sys\_nb1\_na1\_place\_sys\_ner\_locforce\_remove\_status\_loc\_final\_cbia\_urg-self-v4-setS\_IL10\_NN\_CP2\_submissions\_v1.noNI.lemma.dom\_simple.v1.thres.-1.5\_il10\_cp2.tgz}                      \\
il10\_sf\_system18 & constrained   & CP1 & \seqsplit{edl\_output\_v1\_outputs\_il10\_il10\_all\_uniqlast\_place\_sys\_ner\_status\_loc\_cbia\_urg-all\_speech\_v2\_code\_kw\_outputs\_il10\_sin\_setE\_il10\_sin\_setE\_sfs\_KWD1\_k\_2\_th\_0.9\_il10\_cp1.tgz}                                                 \\
il10\_sf\_system18 & constrained   & CP2 & \seqsplit{edl\_output\_v7\_outputs\_il10\_il10\_sys\_nb1\_na1\_place\_sys\_ner\_locforce\_remove\_status\_loc\_final\_cbia\_urg-self-v4-setS\_IL10\_KWD\_CP2\_submissions\_IL10\_sin\_setE\_NI\_only\_sinmorphv4\_speech\_th\_0.8\_il10\_cp2.tgz}                     \\
il10\_sf\_system19 & unconstrained & CP2 & \seqsplit{unconstrained\_edl\_output\_v7\_outputs\_il10\_il10\_sys\_nb1\_na1\_place\_sys\_ner\_locforce\_remove\_status\_loc\_final\_IL10\_NN\_CP2\_submissions\_v1.noNI.ortho.dom\_simple.v1.thres.-1.5\_il10\_cp2.tgz}                                            \\
\bottomrule 
\end{tabular}
\end{table*}

\end{document}